\documentclass[runningheads]{llncs}

\usepackage[mobile]{eccv}

\usepackage{eccvabbrv}

\usepackage{graphicx}
\usepackage{booktabs}

\usepackage[accsupp]{axessibility}  

\usepackage{hyperref}

\usepackage{orcidlink}

\usepackage{threeparttable}
\usepackage[nointegrals]{wasysym}
\usepackage{array}
\usepackage{booktabs}
\usepackage{wrapfig}

\usepackage{multirow}
\usepackage{amsmath}
\newtheorem{myremark}{Remark}

\begin{document}

\title{Regulating Model Reliance on Non-Robust Features by Smoothing Input Marginal Density} 

\titlerunning{Regulating Model Reliance on Non-Robust Features}

\author{Peiyu Yang\inst{1}\orcidlink{0000-0002-3827-8476} \and
Naveed Akhtar\inst{2}\orcidlink{0000-0003-3406-673X} \and
Mubarak Shah\inst{3}\orcidlink{0000-0001-6172-5572} \and
Ajmal Mian\inst{1}\orcidlink{0000-0002-5206-3842
}}

\authorrunning{P.~Yang et al.}

\institute{The University of Western Australia, Perth, Australia\\
\and
The University of Melbourne, Melbourne, Australia\\
\and
University of Central Florida, Orlando, America\\
\email{\ peiyu.yang@research.uwa.edu.au,  naveed.akhtar1@unimelb.edu.au, shah@crcv.ucf.edu, ajmal.mian@uwa.edu.au}}

\maketitle

\begin{abstract}

Trustworthy machine learning necessitates meticulous regulation of model reliance on non-robust features. We propose a framework to delineate and regulate such features by attributing model predictions to the input. Within our approach, robust feature attributions exhibit a certain consistency, while non-robust feature attributions are susceptible to fluctuations. This behavior allows identification of correlation between model reliance on non-robust features and smoothness of marginal density of the input samples. Hence, we uniquely regularize the gradients of the marginal density w.r.t.~the input features for robustness. We also devise an efficient implementation of our regularization\footnote{Our code is available at \url{https://github.com/ypeiyu/input_density_reg}.} to address the potential numerical instability of the underlying optimization process. Moreover, we analytically reveal that, as opposed to our marginal density smoothing, the prevalent input gradient regularization smoothens conditional or joint density of the input, which can cause limited robustness. Our experiments validate the effectiveness of the proposed method, providing clear evidence of its capability to address the feature leakage problem and mitigate spurious correlations. Extensive results further establish that our technique enables the model to exhibit robustness against perturbations in pixel values, input gradients, and density.
\end{abstract}

\section{Introduction}
Research on mitigating model reliance on non-robust input features has recently gained increasing attention due to high-stake machine learning applications \cite{Rudin2019Stop,grathwohl2020your,srinivas2021rethinking,dombrowski2022towards}. In this paper, we advance this direction by introducing a regularization technique that promotes a smooth marginal probability density function of the input to regulate the model's reliance on non-robust features.

To distinguish between robust and non-robust features, we leverage the notion of attributions~\cite{Zeiler2014Visualizing,fong2017interpretable,Sundararajan2017Axiomatic}. For a model $f$ parameterized by $\theta$, attributions characterize the importance of the $i$-th feature $x_i$ of the input $x$ for the model prediction by quantifying the output change between $f(x;\theta)$ and $f(x_{[x_i=0]};\theta)$. Since robust input features contribute to model predictions equally well across slight condition variations, their attributions exhibit a certain consistency. On the other hand, non-robust feature attributions fluctuate under such variations. This potentially identifies a  correlation between the model's reliance on non-robust features and the smoothness of the marginal probability density function of the input samples $p_{\theta}(x)$. This is an important insight for our contribution. For robustness, it offers a possibility of model regularization using the gradients of the marginal density with respect to the input $\nabla_x p_{\theta}(x)$. We propose to regularize $\nabla_x p_{\theta}(x)$ to encourage the model to prioritize the use of robust features and regulate its reliance on non-robust features. However, this can also lead to numerical instability during model optimization. To address that, we further introduce a  stable and efficient implementation to estimate the gradient of marginal density.

\begin{figure*}[t]
	\centerline{\includegraphics[width=1.0\textwidth]{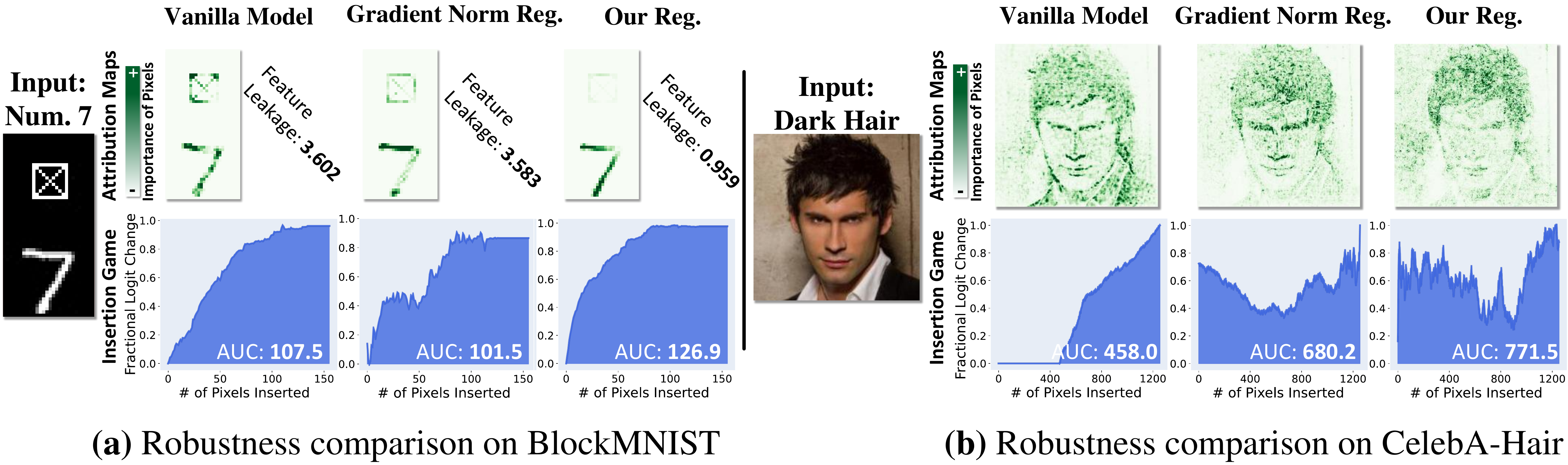}}
	\centering
	\caption{Attribution maps~\cite{Shrikumar2017Learning} and insertion game scores~\cite{Petsiuk2018RISE} for samples from (a) BlockMNIST and (b) CelebA-Hair datasets. As compared to input gradient regularization, our regularization leads to lower feature leakage while also achieving higher AUC for the insertion game.
 }
 \label{fig:intro}
 \vspace{-5mm}
\end{figure*}

We also investigate input gradient norm regularization \cite{Drucker1992Improving,Ross2017Right,Ross2018Improving} 
and reveal that input gradients can be interpreted as input gradients of the log-conditional density $\nabla_x log\ p_{\theta}(x|y)$ or log-joint density $\nabla_x log\ p_{\theta}(x,y)$. Input gradient regularization mitigates the model's reliance on non-robust features specific to the class label $y=i$, leading to unintentional blindness to class-specific non-robust features where $y\neq i$. In contrast, our regularization encourages smoothness of the marginal density $p_{\theta}(x)$ without imposing unintended constraints, providing a comprehensive regularization of the non-robust features. In Fig.~\ref{fig:intro}, 
we employ attribution maps~\cite{Shrikumar2017Learning} and insertion game scores~\cite{Petsiuk2018RISE} to compare the robustness of vanilla models, input gradient regularized models  
and models trained with our regularization on BlockMNIST~\cite{Shah2021Do} and CelebA-Hair~\cite{liu2015faceattributes} datasets. As the representative examples show, our method suppresses both feature leakage and feature spurious correlation to improve model robustness and interpretability. 


The effectiveness of our approach is extensively established through a series of experiments. First, using BlockMNIST dataset~\cite{Shah2021Do}, we demonstrate that the model trained with our regularization considerably mitigates the problem of feature leakage. This problem occurs when a model wrongly attributes importance to irrelevant but persistent features in the data to achieve better accuracy, \eg, the \textit{null} block ($\boxtimes$) in Fig.~\ref{fig:intro}. Addressing feature leakage helps suppress spurious correlation between the input features and model predictions~\cite{Sagawa2019Distributionally,adebayo2020debugging}. 
Moreover, we establish the robustness of the models trained with our regularization against perturbations from adversarial attacks~\cite{Goodfellow2015Explaining,Madry2018Towards}, input pixels~\cite{Samek2017Evaluating} and input density, demonstrating the broad applicability of our approach. Our main contributions are summarized as follows.

\begin{enumerate}
    \item We identify robust and non-robust features by leveraging attributions, and establish the correlation between model reliance on non-robust features and the smoothness of data marginal density.
    \item We propose an efficient technique for calculating the gradients of log density, also addressing the numerical instability of the underlying optimization.
    \item Through extensive experiments, we demonstrate the effectiveness of the proposed regularization and additionally establish that our approach exhibits general robustness against perturbations. 
\end{enumerate}

\section{Related Work}
\noindent{\bf Regularization for Interpretability Robustness:}
Despite advancements in model transparency~\cite{salahuddin2022rtansparency,yang2022multi,jiang2024efficient,jiang2024fast}, current deep neural networks still lack interpretability in their decision-making process, which is exacerbated by their reliance on non-robust input features.
Prior studies, such as~\cite{adebayo2020debugging,Shah2021Do,adebayo2022post}, identify that standard models are prone to relying on irrelevant or spuriously correlated features.
To address that, several regularization techniques are proposed to improve model interpretability. In~\cite{Schramowski2020Making,Erion2021Improving}, the authors incorporated prior knowledge into the model training process to regularize the model behavior. 
Dombrowski \etal~\cite{dombrowski2019explanations,dombrowski2022towards} found that regularizing the input Hessian using SoftPlus activations or weight decay can boost resilience against manipulated inputs. In~\cite{grathwohl2020your}, a joint energy-based model is trained as a discriminative model for improved robustness. 
Srinivas and Fleuret~\cite{srinivas2021rethinking} enhanced the model interpretability by improving the alignment between the implicit density and the data distribution.

\noindent{\bf Regularization for Adversarial Robustness:}
In addition to the other sources of prediction unreliability, adversarial attacks can manipulate model outputs with imperceptible perturbations to inputs~\cite{Goodfellow2015Explaining,Madry2018Towards}. To address this, adversarial training through data augmentation with adversarial samples is widely employed~\cite{carlini2017towards,Madry2018Towards,shafahi2019adversarial}. Certified adversarial robustness through regularizations~\cite{Ross2018Improving,Simon2019First,Anil2019Sorting} is another branch of methods to defend against adversarial perturbations. Inspired by the classic double backpropagation \cite{Drucker1992Improving}, 
Ross and Doshi-Velez~\cite{Ross2018Improving} regularized the input gradient norm for adversarial robustness. Etmann~\cite{etmann2019closer} further explored different variants of double backpropagation regularizations for various real-world scenarios. 
Moosavi-Dezfooli \etal~\cite{moosavi2019robustness} also proposed regularization to encourage a low curvature for adversarial robustness.
To improve adversarial robustness, Chen \etal~\cite{Chen2019Robust} computed the norm of attributions integrated from clean samples to adversarial samples as the regularization term.
Ilyas~\etal~\cite{ilyas2019adversarial} proposed a disentangling method to distinguish feature robustness for explaining adversarial examples. However, they do not focus on general feature robustness. From their  perspective, non-robust features are highly predictive; yet adversarially brittle.
%
In contrast, 
our work focuses on general robustness derived directly from natural images.

\noindent{\bf Attribution Methods:}
Feature attribution methods are used to estimate the importance of input features for a model's prediction and can be categorized as either perturbation-based or back-propagation-based techniques. Perturbation-based methods \cite{Zeiler2014Visualizing,fong2017interpretable,Zintgraf2017Visualizing} calculate the attribution scores by repeatedly perturbing the input features and analyzing the resulting effects on the model prediction. These methods are also extended for evaluating the reliability of the computed attributions \cite{Samek2017Evaluating,Petsiuk2018RISE,Ancona2018Towards,Yang2023Local}.
The back-propagation-based techniques   \cite{Simonyan2014Deep,Sundararajan2017Axiomatic,yang2023recal,yang2024backdoor} estimate the attribution scores by computing the gradients or integrated gradients with respect to the input features in the backward propagation process. 
In contrast to perturbation-based techniques, back-propagation-based attribution methods offer notable advantages in terms of computational efficiency and reliability.
Given the transparency of attributions, our work leverages the attribution framework to distinguish between robust and non-robust features, which enables us to systematically analyze the robustness of input features.

\section{Feature Robustness by Attributions}
We first provide a framework for distinguishing between robust and non-robust features by analyzing their attributions. Herein, attribution inconsistencies among the features with distinct degrees of robustness identify a correlation between the model's reliance on non-robust features and the smoothness of output logits.

Let us consider an input sample $x \in \mathbb R^{n}$ with label $y \in \mathbb R^{c}$ from a dataset $\mathcal D$, and a classifier $f: \mathbb R^n \rightarrow \mathbb R^c$ parameterized by $\theta$. We denote robust and non-robust features within the input $x$ as $x_{rob}, x_{nrob} \subseteq x$. Consider an attribution method $\phi: \mathbb{R}^c \rightarrow \mathbb{R}^n$ attributing model predictions to input features by estimating their importance, resulting in an attribution map $M=\phi(f(x;\theta))$. Inspired by the success of attributions in model explanation, we identify robust and non-robust features by leveraging their attributions.

Without loss of generality, we assume that attributions $M$
of the features can be estimated by calculating the change in output logits when these features are removed from the input, following perturbation-based methods~\cite{Zeiler2014Visualizing,ribeiro2016should}: $M_{x_{rob}} = f(x;\theta) - f(x_{[x_{rob}=0]};\theta)$ and $M_{x_{nrob}} = f(x;\theta) - f(x_{[x_{nrob}=0]};\theta)$. For ease of understanding, we alternatively use $f(x_{nrob};\theta)$ and $f(x_{rob};\theta)$ to represent attributions $M_{rob}$ and $M_{nrob}$ in the text to follow. We define robust features within the attribution framework as follows.

\begin{definition}\label{rob_def}
A feature $x_{feat}$ shared among different input instances under its domain $\Delta_{x_{feat}}$ is robust if, for a randomly chosen class $y=i$, its attribution $M_{x_{feat}}$ is bounded by a small constant $h$ under a metric $c(\cdot)$, i.e.,  $c(f(x_{feat};\theta)-f(\tilde{x}_{feat};\theta)) \leq h : \tilde{x}_{feat} \in \Delta_{x_{feat}}$.
\end{definition}

Under Definition~\ref{rob_def}, robust features are expected to contribute consistently to the model's prediction across different input samples. Non-robust features, on the other hand, are those that contribute to the prediction score inconsistently or only under specific conditions. Our focus here is on distinguishing between robust and non-robust features, without requiring to specify a particular metric. Definition~\ref{rob_def} emphasizes attribution consistency for robust features rather than attribution positivity, thereby allowing for robust features that can also make a negative contribution to the model's prediction. Building further upon the above definition, we make the following remark.

\begin{myremark}
\label{remark1}
Robust features are largely \textbf{condition-invariant} in that they retain similar attributions despite slight changes to the input. In contrast, non-robust features are \textbf{condition-specific} in that their attributions either vary drastically with slightly varying input conditions, or behave robustly only under specific conditions.
\end{myremark}

Robust features exhibit stable behavior across the input space, which is observable through consistent output logits $f(x_{rob};\theta)$ in classifiers regardless of the input instance or class $y$.  
In contrast, non-robust features rely on specific conditions to exhibit a particular behavior, which is tailored to a specific class $y=i$ or the input instance. Robust modeling aims for a stronger reliance on robust features for prediction. Due to the consistency of output logits $f(x;\theta)$ for robust features, a smooth  $f(x;\theta)$ is a desirable property for model robustness.

\section{Smoothing Marginal Density of Input}
Here, we establish the relation between model robustness and gradients of the input marginal density. Then, a robust regularization is proposed for regulating model reliance on non-robust features by smoothing marginal density.

We commence our analytical analysis with probability density, following Bridle~\cite{bridle1990probabilistic}. Given a class $y=i$, a joint probability density function over the input with the output logit $f_i(x;\theta)$ is defined as
\begin{equation} \label{eq_joint}
p_{\theta}(x,y=i) = e^{f_i(x;\theta)} / Z_{\theta},
\end{equation}
where the constant $Z_{\theta}=\int e^{f_i(x;\theta)} dx$ is the partition function.
$Z_{\theta}$ normalizes the input $x$ to a probability density by integrating over all possible input points $x$ in the input space via the model $f$.
By applying Bayes' rule, we eliminate the condition $y=i$, resulting in the marginal density being defined solely on the input $x$: $p_{\theta}(x) = p_{\theta}(y=i, x)/p_{\theta}(y=i| x)$. The conditional density function $p_{\theta}(y=i|x)$ can be further defined as
\begin{equation} \label{eq_cond_pyx}
p_{\theta}(y=i|x) = e^{f_i(x;\theta)} / Z_{f(x;\theta)},
\end{equation}
where $Z_{f(x;\theta)} = \sum^C_{i=1}e^{f_i(x;\theta)}$ is the partition function for the  output logits $f_i(x;\theta)$ defined on all the $C$ classes. To simplify the notation, we use $Z_{f(x)}$ to represent $Z_{f(x;\theta)}$ in the subsequent discussion.
Exploiting the symmetry property of the joint density defined in Eq.~\ref{eq_joint}, i.e., $p_{\theta}(x,y=i) = p_{\theta}(y=i, x)$, the marginal density $p_{\theta}(x)$ can be expressed as
\begin{equation}\label{eq_px}
p_{\theta}(x) = \frac{e^{f_i(x;\theta)}/Z_{\theta}}{e^{f_i(x;\theta)}/Z_{f(x)}
} \\
=\frac{Z_{f(x)}}{Z_{\theta}}.
\end{equation}

As identified in the previous section, smoothness of output logits across inputs encourages the use of robust features by the model. Hence, we consider the marginal density $p_{\theta}(x)$ defined on the output logits $f(x;\theta)$ across the input space. Promoting a smaller gradient of the marginal density with respect to the input $x$, denoted as $\nabla_x p_{\theta}(x)$, contributes to the smoothness of the output logits. Thus, a positive correlation can be established between the use of robust features and the smoothness of the probability marginal density $p_{\theta}(x)$. In particular, the smooth output logits of robust features across input samples suggest that these features will have relatively small gradients of the density $p_{\theta}(x)$ with respect to the input values. On the other hand, non-robust features with fluctuating output logits will have large gradients of the density that need to be suppressed during the training process for model robustness. Therefore, we can conclude with the following remark.

\begin{myremark}
\label{remark2}
Model reliance on non-robust features $x_{nrob}$ can be regulated by regularizing the gradients of the marginal density $p_{\theta}(x)$ w.r.t.~$x$, and this regularization can be achieved through optimizing the model parameters. 
\end{myremark}

In the light of Remark~\ref{remark2}, we propose a regularization term for minimizing the gradients of marginal density. However, computing the gradients of the marginal density $\nabla_x p_{\theta}(x)$ is not feasible because the partition function defined on the entire input space is intractable. To avoid the estimation of the intractable $Z_{\theta}$, we instead compute the gradients with respect to the log density as $\nabla_x log\ p_{\theta}(x) = \nabla_x log\ Z_{f(x)}$. This is possible because $Z_{\theta}$ solely depends on the model parameter $\theta$ and not the input $x$. Expanding the partition function $Z_{f(x)}$ in $\nabla_x log\ p_{\theta}(x)$, we obtain $\nabla_x log\ p_{\theta}(x) = \sum^C_{i=1}\nabla_x e^{f_i(x;\theta)} / \sum^C_{i=1}e^{f_i(x;\theta)}$. The $p$-norm of this gradient is computed as the regularization term. In the optimization process, the goal is to find the optimal parameter $\theta^*$, \textit{cf.} Remark~\ref{remark2}, by minimizing the loss $\ell$ as
\begin{equation} \label{eq_px_grad_2}
\theta^* = \underset{\theta}{\mathrm{min}} \; \ell(f(x;\theta), y) + \lambda ||\frac{\sum^C_{i=1}\nabla_x e^{f_i(x;\theta)}}{\sum^C_{i=1}e^{f_i(x;\theta)}}||_p,
\end{equation}
where $\lambda$ indicates the magnitude of the coefficient for controlling the strength of the regularization.

To regulate model reliance on non-robust features, our regularization encourages the smoothness of marginal density by regularizing its gradients. Since the output logit change in the log-marginal density $log\ p_{\theta}(x)=log\ Z_{f(x)} - log\ Z_{\theta}$, and $Z_\theta$ is independent of the input $x$, we can only focus on the first term $Z_{f(x)} = \sum_{i=1}^C e^{f_i(x;\theta)}$. Recall, Definition~\ref{rob_def} of robust features. Assuming the robust input feature $x_{rob}$ exists in a random input $x$, the corresponding output logit $f_i(x_{rob})$ will consistently attribute to the model prediction. 
This property of $x_{rob}$ leads to the smoothness of output change $\sum_{i=1}^C e^{f_i(x_{rob};\theta)}$ across different input samples and class labels. 
In contrast, non-robust input features $x_{nrob}$ show relatively high attributions for the output logit $f_i(x_{nrob})$ for a given class $i$. However, they cannot maintain consistency in the attributions across different inputs or labels, leading to fluctuations in the output change $\sum_{i=1}^C e^{f_i(x_{nrob};\theta)}$. In Fig.~\ref{fig:intro}(a) and Fig.~\ref{fig:attr_bm} in Supp.~\ref{app:sec_ins_vis}, non-robust features in the \textit{Null} block exhibit fluctuating attributions across different samples on the standard trained model. It is demonstrated that the magnitude of gradients for input features in the marginal density $p_{\theta}(x)$ reflects the model's sensitivity to those features. We leverage this relation to mitigate model reliance on the non-robust features by smoothing the marginal density of the input samples.

\section{Stable and Efficient Implementation for Regularization}
\begin{figure*}[b] 
	\centering
        \vspace{-5mm}
	\includegraphics[width=.85\textwidth]{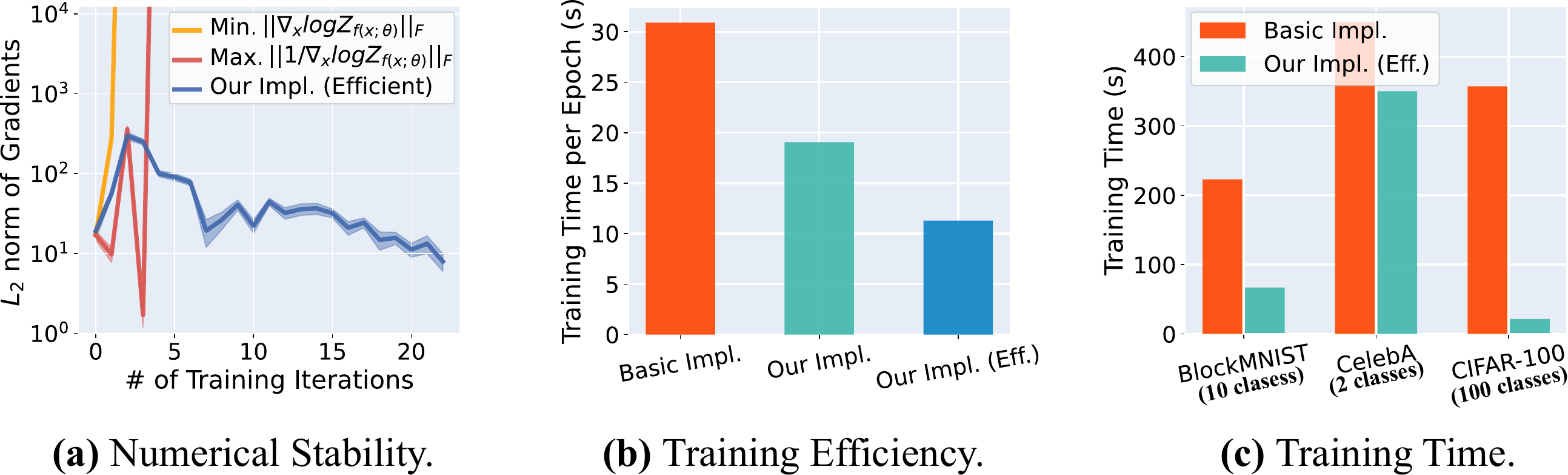}
	\caption{The comparison of numerical stability, training efficiency and training time across different ResNet-34 models.}
	\label{fig:num_stable}
\end{figure*}
From the implementation perspective, the gradient computation of marginal density involves multiple exponential operations in both the numerator and denominator of Eq.~\ref{eq_px_grad_2} which can introduce numerical instability in the optimization process, leading to gradient vanishing and explosion problems. Such issues can potentially hinder the application of our regularization to large non-linear models or wide-distribution data. For instance, batch normalization (BN) layers~\cite{ioffe2015batch} solving internal covariate shifts with learnable scaling and shifting parameters can amplify the errors caused by the exponential operations during the backpropagation.
In Fig.~\ref{fig:num_stable}(a), yellow and red curves indicate original and inverse implementations for minimizing the gradient of log density $\nabla_xlogZ_{f(x)}$. It is illustrated that the Frobenius gradient norm of both implementations undergoes rapid numerical overflow as the number of training iterations increases, revealing the serious issue of exploding and vanishing gradients using the proposed regularization. Therefore, it is crucial to address these through careful implementation to ensure the feasibility of our regularization.

To address this challenge, we transform the computation from the summation of exponential operations to softmax. By subtracting a constant value $\eta$ from the logits before exponentiation, softmax can prevent numerical overflow during the exponential operations, i.e., $e^{f_i(x;\theta)}/\sum_j e^{f_j(x;\theta)} = e^{f_i(x;\theta)-\eta}/\sum_j e^{f_j(x;\theta)-\eta}$. Thus, we incorporate the softmax function into the computation of log density gradient $\nabla_x log z_{f(x)}$ as
\begin{equation} \label{eq_logsoft}
\begin{split}
\nabla_x log z_{f(x)}
\!&= \!\nabla_x f_i(x;\theta) \! - \! \nabla_x (log e^{f_i(x;\theta)} - log z_{f(x)}), \\
\!&= \!\nabla_x f_i(x;\theta) \!-\! \nabla_x (log (f_i(x;\theta)/z_{f(x)})), 
\end{split}
\end{equation}
where $f_i(x)$ indicates a logit of a random $i$-th class, and $z_{f(x)}$ equals $\sum_j e^{f_j(x;\theta)}$. Thus, the gradients with respect to the log-marginal density $p_{\theta}(x)$ can be replaced by computing the difference between the gradient $\nabla_x f_i(x;\theta)$ and the gradient of a log-softmax output $\nabla_x log(e^{f_i(x;\theta)}/Z_{f(x)})$.

Our technique improves upon the common approach~\cite{bridle1989training,blanchard2021accurately} to achieve numerical stability in log exponential sum calculations, which typically employs the formula $log(\sum_{i\in\{1,\dots,n\}} e^{x_i}) =\eta+log(\sum_{i\in\{1,\dots,n\}} e^{x_i-\eta})$, with $\eta$ being the maximum value of inputs $\{x_1,\dots,x_n\}$. We address the numerical instability by employing softmax, avoiding computationally expensive comparisons of the maximum gradient values. Specifically, the basic stable implementation involves finding the maximum gradient within the gradients of various classes, leading to a time complexity of $O(n)$, where $n$ represents the number of classes. In contrast, our method eliminates numerical instabilities associated with a randomly selected class, eliminating the need for maximum value comparisons and leading to a more favorable time complexity of $O(1)$.

While our solution helps avoid numerical instability, it requires twice the gradient computations as compared to the sole calculation of density gradients. Hence, we propose an efficient mechanism for estimating the difference in the gradients. Specifically, we compute the gradient of the difference between two outputs to approximate the difference between the two gradients of outputs using Taylor series as
\begin{equation} \label{eq_taylor}
\nabla_x f_i(x;\theta) - \nabla_x log(e^{f_i(x;\theta)} / Z_{f(x)}) \approx 
\nabla_x (f_i(x;\theta) - log(e^{f_i(x;\theta)} / Z_{f(x)})).
\end{equation}

The proof of Eq.~\ref{eq_taylor} is provided in Supp.~\ref{app:sec_proof}. As such, the proposed approach enables stable and efficient model optimization. The blue curve in Fig.~\ref{fig:num_stable}(a) shows that our efficient implementation can effectively avoid numerical instability in the gradient computation. Fig.~\ref{fig:num_stable}(b) provides a comparison of training efficiency on CIFAR-10~\cite{krizhevsky2009learning} between the basic numerically stable implementation and our two derived alternatives. More results regarding the comparison of numerical stability are provided in Supp.~\ref{app:exp_numerical}.
In Fig.~\ref{fig:num_stable}(c), we delve into a specific examination of the training times of models trained across the three datasets including BlockMNIST, CelebA and CIFAR-10.
It is evident that our methods substantially improve the per-epoch training time. Moreover, our methods can yield further beneficial trade-offs as the number of classes increases.


\section{Limited Robustness in Input Gradient Regularization}
Input gradient regularization (InputGrad Reg.)~\cite{Lecun1998Gradient,Ross2018Improving} computes the Frobenius norm of input gradients $||\nabla_x f_i(x;\theta)||_F$ for a given class label $y=i$, which is a baseline robust regularization for model optimization.
Existing works~\cite{Lecun1998Gradient,Ross2017Right,Ross2018Improving} explain InputGrad Reg.~as a prediction stability technique for robustness against perturbations.
Here, we reveal that the input gradient norm potentially regularizes the gradients of implicit data density. Moreover, we provide an understanding of how InputGrad Reg. encourages robustness as well as its limitations.

Suppose all the classes have equal probability, i.e., $p_{\theta}(y=i) = 1/C$. We can express the class-conditional density $p_{\theta}(x|y)$ by using Bayes's rule as
\begin{equation} \label{eq_cond_pxy}
p_{\theta}(x|y=i) = \frac{p_{\theta}(x,y=i)}{p_{\theta}(y=i)} =\frac{e^{f_i(x;\theta)}}{Z_{\theta}/C}.
\end{equation}

Now, we compute the gradients of the log density defined in Eq.~\ref{eq_cond_pxy} with respect to the input $x$ as
\begin{equation} \label{eq_grad_reg}
\nabla_x log\ p_{\theta}(x,y=i) = \nabla_x log\ p_{\theta}(x|y=i) = \nabla_x f_i(x;\theta).
\end{equation}

Eq.~\ref{eq_grad_reg} demonstrates that the input gradients can be interpreted as the gradients of either the log joint density or the log conditional density with respect to the input $x$. 
This formulation highlights that InputGrad Reg.~encourages consistent attributions of input features for the model's predictions. However, it is important to note that InputGrad Reg.~is formulated under the specific condition $y=i$, which limits its effectiveness in resolving inconsistencies when predicting a different class $y=j$, where $j\neq i$. Although input features consistently contribute to the model's prediction under a given class, InputGrad Reg.~fails to consider inconsistent attributions of these features across different classes, thereby allowing model non-robust behavior to exist. Consequently, a model trained with input gradient regularization may still exhibit spurious robustness relying on specific conditions. Fig.~\ref{fig:intro}(a) illustrates an example where the model trained with InputGrad Reg. is incapable of suppressing non-robust features in \textit{Null} block.

\begin{myremark}
\label{remark3}
Input gradient regularization smoothens the joint and conditional density of the input $x$ under a specific label $y=i$, compromising its ability to resist the class-specific non-robust features.
\end{myremark}

In contrast to the existing works~\cite{Ross2018Improving,srinivas2021rethinking} that highlight the reasons for InputGrad Reg.~efficacy, we reveal a weakness of this technique, \textit{cf.}~Remark~\ref{remark3}. Unlike regularizing gradients based on joint or conditional densities, our approach allows for more effective regularization without imposing the condition $y=i$. Our method focuses on regularizing the gradients of marginal density $\nabla_x p_{\theta}(x)$, thereby smoothing the output logits across the input samples. 

\section{Experiments}
In this section, we perform extensive experiments to validate the efficacy of our regularization and the newly established correlation between the smooth marginal density and model reliance on non-robust features. Specifically, we present measurement results for addressing feature leakage and mitigating spurious correlations.
In addition to specific applications, further results on the robustness against perturbations in input pixels, gradients and density are presented. Additional details about the datasets and the models used in our experiments can be found in Supp.~\ref{app:sec_exp_setup}.

\subsection{Efficacy against Feature Leakage and Adversarial Attacks}
\begin{wrapfigure}{r}{.53\textwidth}
\vspace{-6mm}
	\includegraphics[width=.52\textwidth]{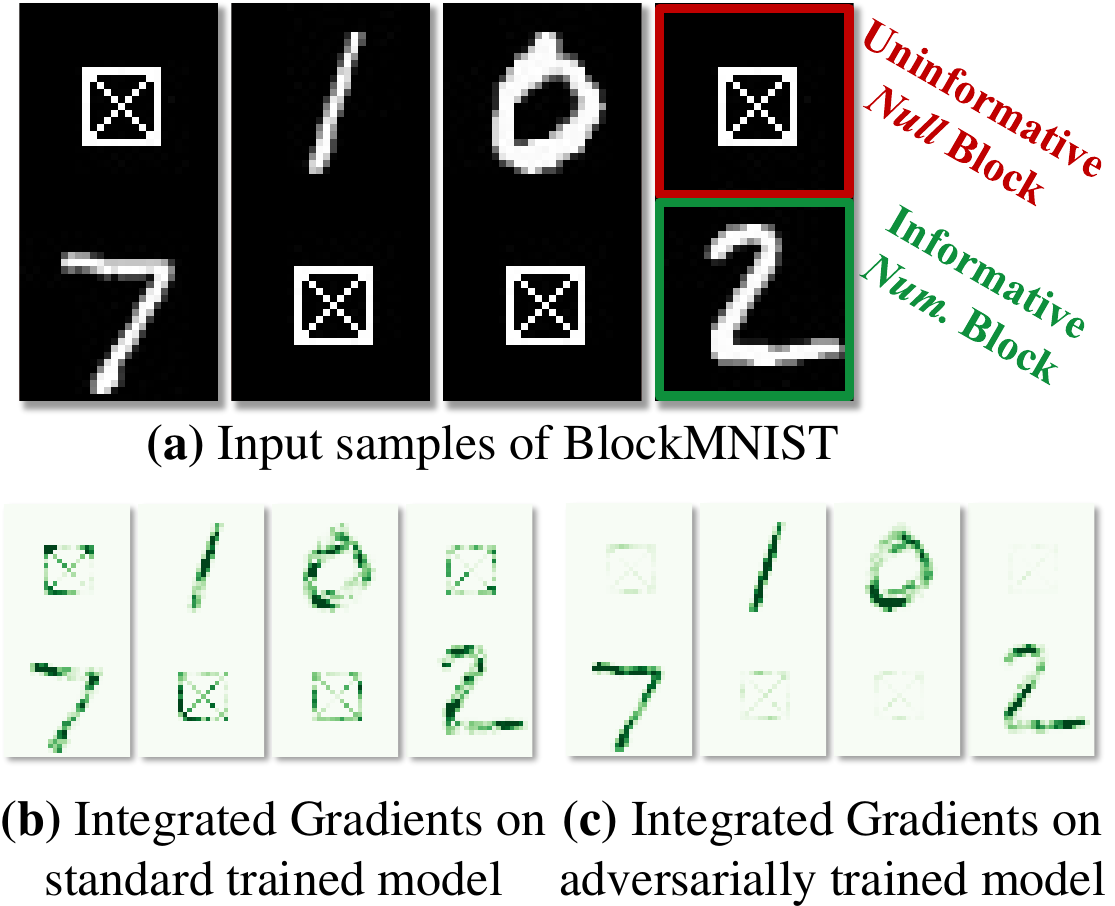}
	\caption{BlockMNIST samples and feature leakage problem. \textbf{(a)} BlockMNIST randomly appends a null block at the top or bottom of MNIST samples. \textbf{(b\&c)} Attribution maps are calculated by IG on the standard and adversarially trained models.}
	\label{fig:BMnist}
\vspace{-4mm}
\end{wrapfigure}

In \cite{Shah2021Do,adebayo2020debugging}, it is demonstrated that deep models end up assigning importance to irrelevant input features.
Shah \etal~\cite{Shah2021Do} used BlockMNIST in their experiments, which is a synthetic dataset extended from MNIST~\cite{Lecun1998Gradient}.
To each MNIST sample, BlockMNIST attaches a \textit{null} block (an irrelevant pattern) randomly at the top or bottom of the image, as shown in Fig.~\ref{fig:BMnist}(a). 
Shah \etal~\cite{Shah2021Do} observed that the explaining tool InputGrad~\cite{Simonyan2014Deep} attributes importance to both the informative number block and the uninformative null block in the standard trained model. This phenomenon is termed as \textit{feature leakage} by the authors. Interestingly, the issue is mitigated in adversarially trained models.
Since this dataset allows for a controlled robustness assessment, we first evaluate our method on BlockMNIST to analyze the model's reliance on irrelevant features.

\noindent\textbf{Reproducibility and Quantitative Measurement of Feature Leakage.}
Owing to the unreliability of InputGrad caused by model saturation~\cite{Shrikumar2017Learning}, we employ Integrated Gradients (IG)~\cite{Sundararajan2017Axiomatic}, an axiomatic explanation tool, to faithfully re-investigate the feature leakage phenomenon. In Fig.~\ref{fig:BMnist}(b)-(c), we show that the attributions computed by IG can reproduce different leakage phenomena from the informative block region to the \textit{null} block region on standard and adversarially trained models. Feature leakage is an important phenomenon in the context of model robustness. However, there is a lack of a quantitative metric in the current literature to quantify its extent. We use integrated gradients to define the metric $M_{leakage}$ to address this gap. Mathematically,
\begin{equation} \label{eq:opt_wass}
M_{leakage} = \underset{x_{nrob} \sim D}{\mathbb{E}} \ ||x_{nrob} \times \int_{\alpha=0}^1 \frac{\partial f(\alpha\cdot x_{nrob}; \theta))}{\partial x_{nrob}} \mathrm{d}\alpha||_2,
\end{equation}
where $\alpha$ is the step from the absence to the presence of the features, and $x_{nrob}$ specifies the non-robust features in the null block. Since the attributions of $x_{nrob}$ for the model's prediction should ideally be zero, we use the $L_2$ norm of attributions to quantify the extent of feature leakage.

\noindent\textbf{Robustness against  Feature Leakage.} Table~\ref{table:robust_BM} presents the experimental results on the BlockMNIST dataset. We compare our method with other robust regularizations and techniques including InputGrad~\cite{Ross2018Improving}, IG-SUM~\cite{Chen2019Robust}, SoftPlus activations~\cite{dombrowski2019explanations} and Hessian~\cite{dombrowski2022towards}.
InputGrad and Hessian regularize the first-order and second-order gradients w.r.t.~the input. Models trained with SoftPlus activations and Hessian regularization fail to suppress the leakage problem, which indicates that feature leakage is not caused by the geometry of the model output manifold or high curvature~\cite{dombrowski2019explanations,zhang2020interpretable}. InputGrad regularization demonstrates robustness against both $L_{2}$ and $L_{\infty}$ adversarial attacks, yet it still fails to address the leakage problem.
The result aligns well with Remark~\ref{remark3} which highlights the allowance of non-robust features across different classes in InputGrad regularization.
In addition, the use of IG in IG-Norm regularization that accumulates input gradients as a regularization term results in behavior similar to InputGrad regularization, leading to compromised robustness.
These results further reveal that adversarial robustness is not a sufficient condition for suppressing feature leakage. Our method demonstrates a considerable improvement over other techniques for feature leakage, while also maintaining superiority in adversarial robustness. In Tab.~\ref{table:robust_BM}, we use $L_2$ norm for the compared regularization terms for fair benchmarking.
We show in Supp.~\ref{app:sec_norm} that optimal norm exploration can yield an even more favorable trade-off for our technique for model robustness.

%
\begin{table}[b]
\vspace{-3mm}
\caption{Results on BlockMNIST. Feature leakage, standard accuracy and adversarial accuracy under $L_2$ and $L_{\infty}$ PGD-20 attacks are reported. ST $\&$ AT: Standard and Adversarial Training.}
\centering
\label{table:robust_BM}
\setlength{\tabcolsep}{1.6mm}{
\begin{tabular}{lccccc}
\toprule[1.5pt]
Method & Feature Leakage $\downarrow$ & PGD-20 ($L_{2}$) $\uparrow$ & PGD-20 ($L_{\infty}$) $\uparrow$ & Accu. $\uparrow$ \\
\hline
AT (FGSM) & 3.324 & 87.48 & 0.00 & 99.02\\
AT (PGD) & 2.313 & 92.75 & 28.05 & 98.97\\
\hline
ST & 3.657 & 73.57 & 0.00 & \textbf{99.12}\\
+ SoftPlus Acts. & 3.533 & 67.95 & 0.02 & 98.52\\
+ Hessian Reg. & 4.258 & 80.06 & 0.00 & 98.48\\
+ InputGrad Reg. & 3.461 & 83.46 & 21.14 & 94.56\\
+ IG-SUM Reg. & 3.497 & 82.11 & 22.73 & 92.87\\
+ \textbf{Our Reg.} & \textbf{2.259} & \textbf{85.41} & \textbf{29.36} & 93.05\\
\bottomrule[1.5pt]
\end{tabular}
}
\vspace{-5mm}
\end{table}

\noindent\textbf{Feature Leakage in Adversarially Robust Models.} A FGSM adversarially trained model~\cite{Goodfellow2015Explaining} augments the training samples by adversarial examples $x+\epsilon \cdot sign(\nabla_xf_i(x;\theta))$.
Notably minimizing the loss of the perturbed input $x+\epsilon \cdot sign(\nabla_xf_i(x;\theta))$ is similar to the InputGrad regularization. Thus, training with FGSM is still limited in its ability to suppress the leakage problem. In contrast, PGD attack~\cite{Madry2018Towards} weakens the effect of the condition $y=i$ by iteratively searching for the perturbations from a random starting point. This process leads to a substantial enhancement in suppressing feature leakage, see Tab.~\ref{table:robust_BM}.%

\begin{figure*}[t]
    \begin{subfigure}[c]{0.23\textwidth}
        \centering
        \includegraphics[width=\textwidth]{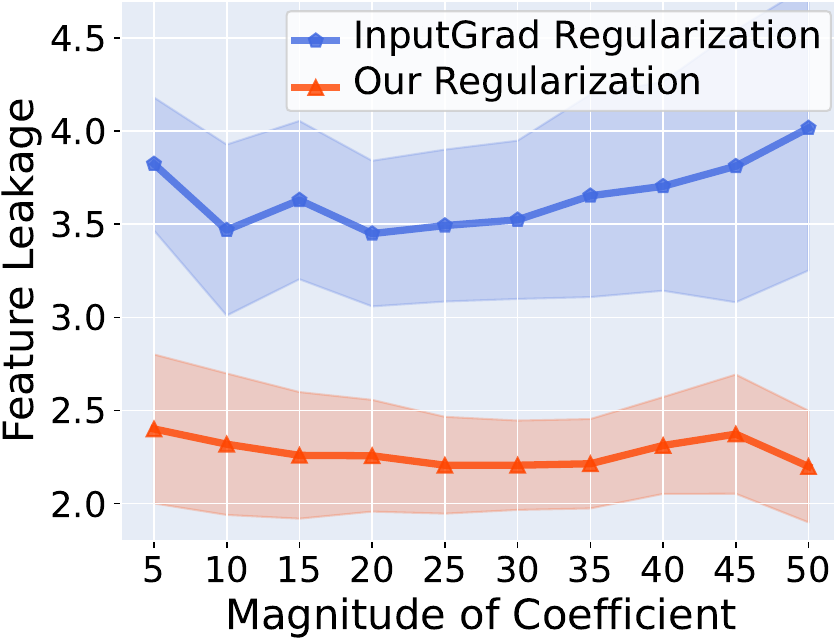}
        \caption{Feature Leakage.}
        \label{fig:leak}
    \end{subfigure}
    \hfill
    \begin{subfigure}[c]{0.23\textwidth}
        \centering
        \includegraphics[width=\textwidth]{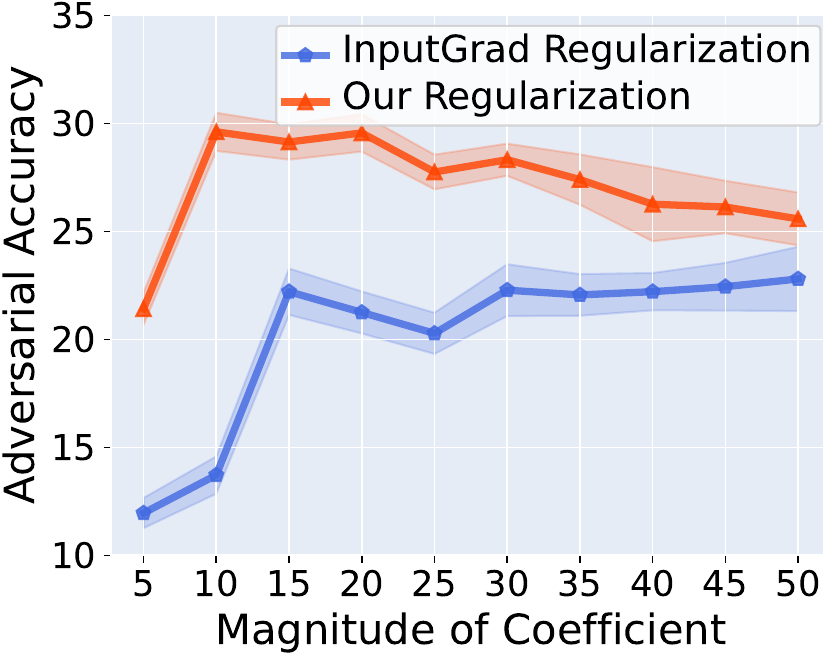}
        \caption{$L_{\infty}$ Attack.}
        \label{fig:adv_lin}
    \end{subfigure}
    \hfill
    \begin{subfigure}[c]{0.23\textwidth}
        \centering
        \includegraphics[width=\textwidth]{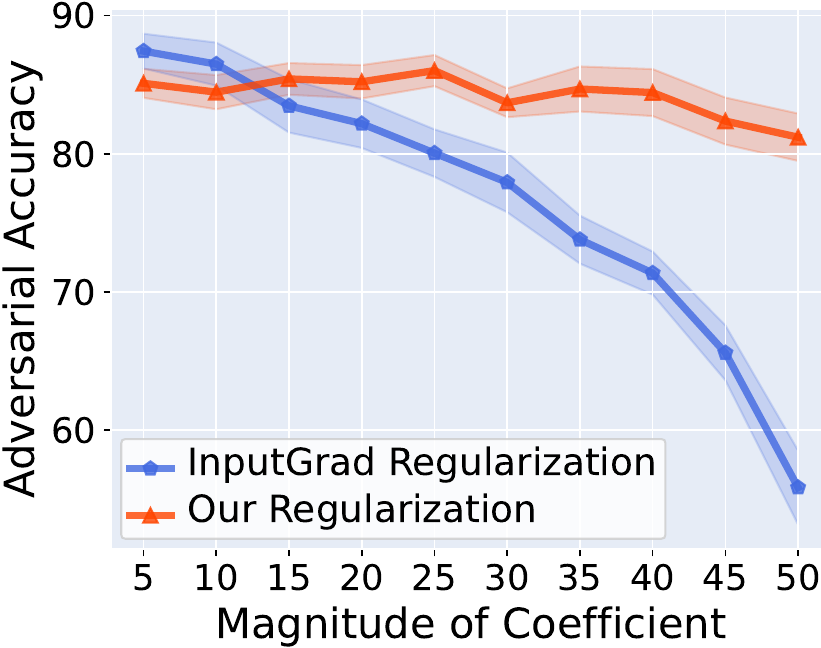}
        \caption{$L_{2}$ Attack.}
        \label{fig:adv_l2}
    \end{subfigure}
    \hfill
    \begin{subfigure}[c]{0.23\textwidth}
        \centering
        \includegraphics[width=\textwidth]{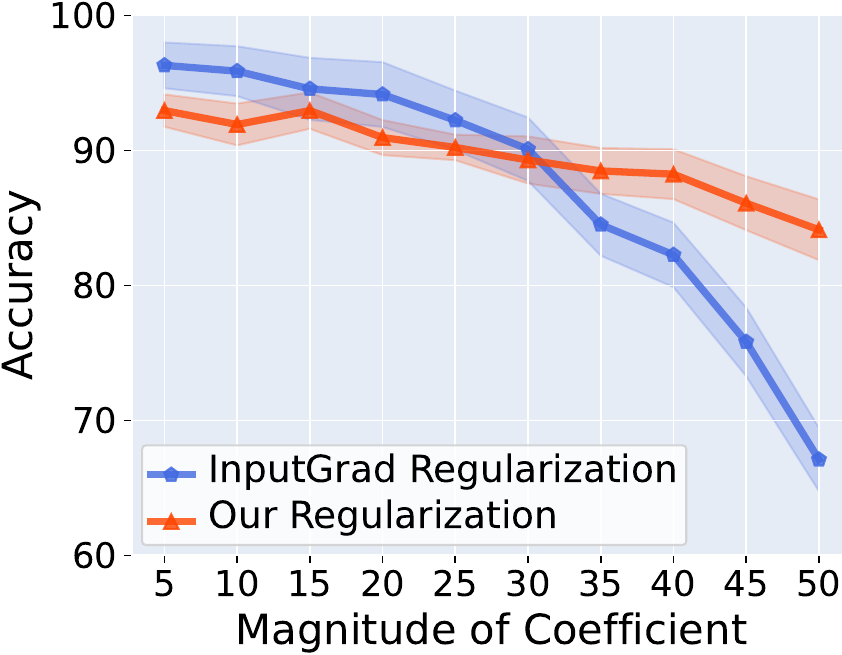}
        \caption{Accuracy.}
        \label{fig:acc}
    \end{subfigure}
    \caption{Performance comparison between our method and InputGrad regularization under varying regularization coefficient for (a) Feature leakage, (b)-(c) Adversarial Accuracy under $L_{\infty}$ and $L_{2}$ PGD-20 attacks, and (d) Accuracy. }
    \vspace{-5mm}
    \label{fig:robust_bm}
\end{figure*}

\noindent\textbf{Magnitude of Coefficient for Regularization.} 
Figs.~\ref{fig:robust_bm}(a)-(d) present a comparison of results for feature leakage and adversarial accuracy under PGD attacks, as well as the standard accuracy across varying magnitudes for the regularization strength.
The results affirm that our method effectively regulates feature leakage by imposing a penalty on non-robust features. Moreover, our regularization enables the model to defend against both $L_2$ and $L_{\infty}$ attacks while maintaining high accuracy, showing an outstanding trade-off across four criteria. More adversarial robustness comparisons on CIFAR dataset~\cite{krizhevsky2009learning} are reported in Supp.~\ref{app:sec_adv_rob}.

\subsection{Efficacy for Spurious Correlation}
\begin{wrapfigure}{r}{.5\textwidth}
\vspace{-6mm}
	\includegraphics[width=.5\textwidth]{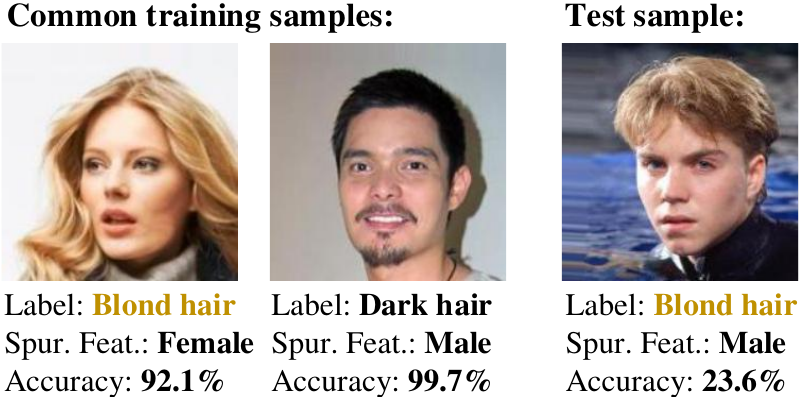}
	\caption{Spurious correlation on ResNet-34 trained on CelebA-Hair. The model fails to classify the male celebrity with blond hair due to a spurious correlation learned between females and blond hair.}
	\label{fig:celeba}
\vspace{-5mm}
\end{wrapfigure}
Recent research has highlighted the susceptibility of neural models in learning spurious correlations that enhance performance on a given data but fail to generalize~\cite{buolamwini2018gender,Sagawa2019Distributionally,adebayo2020debugging}. For instance, in CelebA-Hair dataset~\cite{liu2015faceattributes}, which commonly consists of samples containing female celebrities with blond hair and male celebrities with dark hair, models heavily rely on the spuriously correlated \textit{gender} feature to predict the target \textit{hair color}~\cite{hashimoto2018fairness,Sagawa2019Distributionally}. Consequently, accuracy tends to be lower for the samples containing male celebrities with blond hair, see Fig.~\ref{fig:celeba}.

To mitigate spurious correlations, distributional robust optimization (DRO) techniques have been proposed to re-weight the training loss of input samples from different groups~\cite{hu2018does,Sagawa2019Distributionally}. In our regularization, an additional penalty is imposed to penalize the model's reliance on these spuriously correlated features because of their inconsistent attributions. This encourages the use of robust features while suppressing the model's reliance on spuriously correlated features. Reliable quantification of the model's robustness on natural images for spurious correlation suppression is an unresolved issue in the literature. We employ attribution maps~\cite{Sundararajan2017Axiomatic} and insertion game~\cite{Petsiuk2018RISE} to demonstrate the effectiveness of our regularization in suppressing the use of the spuriously correlated \textit{gender} feature and promoting the use of the robust \textit{hair} feature, as shown in Fig.~\ref{fig:intro}(b). This superiority leads to performance improvements on worst-case samples in the model trained with our regularization.

\begin{table*}[t]
    \centering
    \caption{Worst-group accuracy and overall accuracy comparisons between Vanilla ResNet-34, group DRO, Score-Matching, InputGrad and our regularization on CelebA-Hair and Waterbirds datasets.}
    \label{table:tab_DRO}
    \setlength{\tabcolsep}{1.8mm}{
    \begin{tabular}{lcccc}
        \toprule[1.5pt]
        \multirow{2}{*}{Method} & 
        \multicolumn{2}{c}{Worst-Group Accuracy (\%)} &
        \multicolumn{2}{c}{Overall Accuracy (\%)} \\ \cmidrule(r){2-3} \cmidrule(r){4-5} & 
        CelebA-Hair & Waterbirds & CelebA-Hair & Waterbirds \\
        \hline
        Vanilla Model & $49.90_{\ \pm \ 8.69}$ & $62.90_{\ \pm \ 0.10}$ & ${94.90}_{\ \pm \ 0.39}$ & ${87.70}_{\ \pm \ 0.08}$ \\
        Group DRO & $59.44_{\ \pm \ 5.98}$ & $63.60_{\ \pm \ 0.17}$ & $\boldsymbol{94.96}_{\ \pm \ 0.21}$ & $87.60_{\ \pm \ 0.05}$ \\        
        Score-Matching Reg. & $59.78_{\ \pm \ 7.56}$ & $58.19_{\ \pm \ 1.55}$ & $93.46_{\ \pm \ 0.71}$ & $85.64_{\ \pm \ 0.38}$ \\
        InputGrad Reg. & $82.66_{\ \pm \ 3.63}$ & $58.18_{\ \pm \ 1.22}$ & $92.12_{\ \pm \ 2.18}$ & $85.50_{\ \pm \ 0.27}$ \\
        \textbf{Our Reg.} & $\boldsymbol{85.62}_{\ \pm \ 5.36}$ & ${63.78}_{\ \pm \ 2.83}$ & $92.30_{\ \pm \ 1.38}$ & $86.48_{\ \pm \ 0.38}$ \\
        \textbf{Group DRO + Ours} & ${82.98}_{\ \pm \ 4.69}$ & $\boldsymbol{73.82}_{\ \pm \ 2.27}$ & $93.62_{\ \pm \ 0.74}$ & $\boldsymbol{90.52}_{\ \pm \ 0.17}$ \\
        \bottomrule[1.5pt]
    \vspace{-5mm}
    \end{tabular}}
\end{table*}

Table~\ref{table:tab_DRO} presents comparison of accuracy on worst-group samples and overall samples from the CelebA-Hair and Waterbirds datasets~\cite{liu2015faceattributes,Sagawa2019Distributionally}. The Waterbirds dataset consists of synthetic bird images from CUB-200-2011~\cite{wah2011caltech} and Places~\cite{zhou2017places} datasets,  incorporating spurious background features, such as land and water scenes, to confuse true labels of bird categories.
We compare the proposed regularization method with InputGrad and Score-Matching regularizations~\cite{srinivas2021rethinking}, as well as Group DRO~\cite{Sagawa2019Distributionally}. Score-Matching regularization is proposed to enhance the interpretability of the model by improving the alignment of implicit density models.
Experimental results clearly demonstrate the effectiveness of our method in enhancing worst-group accuracy while maintaining overall sample accuracy. Furthermore, the performance gains can be further enhanced by incorporating our regularization technique into Group DRO. This enhancement highlights the efficacy of our regularization in terms of its practicality. 
More results using attribution maps and the insertion game on both CelebA-Hair and Waterbirds datasets are provided in Supp.~\ref{app:sec_ins_vis}. 
Moreover, out-of-distribution detection~\cite{hendrycks2016baseline} is also performed on CIFAR~\cite{krizhevsky2009learning} and SVHN~\cite{netzer2011reading} in Supp.~\ref{app:sec_ood}.

\subsection{Efficacy against Pixels, Gradients and Density Perturbations}
\begin{figure*}[t]
\centerline{\includegraphics[width=1.0\textwidth]{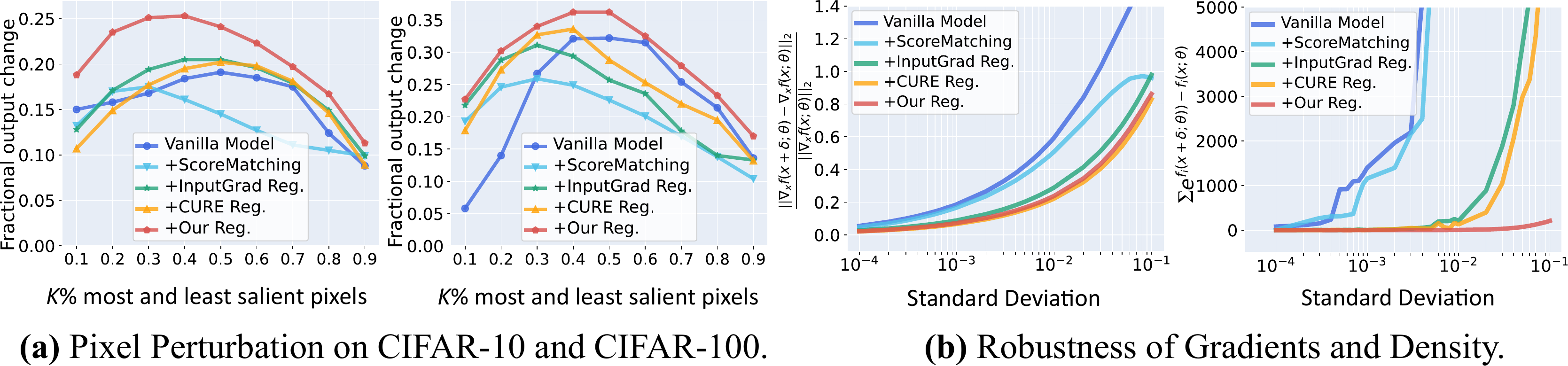}}
	\centering
	\caption{Robustness comparison. \textbf{(a)} Pixel perturbation experimental results on ResNet-18 trained on CIFAR-10 and CIFAR-100. {Higher} curves indicate {better} results. \textbf{(b)} Robustness of relative gradients and absolute density on CIFAR-10. {Lower} curves indicate {better} results. }
     \vspace{-3mm}
	\label{fig:pert_rob}
\end{figure*}

We first employ pixel perturbation~\cite{Samek2017Evaluating,Yang2023Local} to quantitatively compare the robustness of different models following
Srinivas and Fleuret~\cite{srinivas2021rethinking} who iteratively removed the most important input pixels identified by attribution maps for model robustness evaluation. Robust models are expected to exhibit increased sensitivity when removing the most important pixels and decreased sensitivity when removing the least important ones. We assess the difference in fractional output logit change between the images with the top and bottom $k$\% most salient pixels using SmoothGrad~\cite{Smilkov2017Smoothgrad} on ResNet-18~\cite{He2016Identity} trained on CIFAR-10 and CIFAR-100~\cite{krizhevsky2009learning}, as depicted in Fig.~\ref{fig:pert_rob}(a).
Given the recognized challenge of reference image ambiguity in IG for explaining natural images~\cite{Erion2021Improving,Sturmfels2020Visualizing}, we opt to use SmoothGrad for explaining predictions of images in the CIFAR dataset. SmoothGrad has demonstrated to outperform IG in such conditions~\cite{Hooker2019Benchmark,Srinivas2019Full}. Recognizing their respective strengths, we use different attribution methods in appropriate conditions for reliable results.
It can be observed that models trained with our regularization significantly outperform different robust regularizations including Score-Matching, InputGrad and CURE regularizations~\cite{moosavi2019robustness}.

We also test the robustness of the relative input gradients 
$||\nabla_x f(x+\delta)-\nabla_x f(x)||_2 / ||\nabla_x f(x)||_2$
and the absolute density 
through $\sum^C_{i=1} e^{f_i(x+\delta)-f_i(x)}$
on the input with Gaussian noise $\delta$ in increasing standard deviation on CIFAR-10 dataset, as shown in Fig.~\ref{fig:pert_rob}(b). We can observe that our regularization leads to competitive robustness for the relative gradient in comparison with regularizing the Hessian norm in CURE. Moreover, our regularization naturally leads to density robustness, which is associated with a strong generative ability of the models~\cite{srinivas2021rethinking}. More tests on CIFAR-100 are provided in Supp.~\ref{app:sec_pert_rob}. Supp.~\ref{app:sec_act_vis} provides the visualizations calculated by activation optimization. These results affirm that our regularization improves both the discriminative and generative abilities of models.

\section{Conclusion}

In this paper, we define robust and non-robust features from a feature attribution perspective, and establish a correlation between the smoothness of input marginal density and model reliance on non-robust features. This connection motivates us to propose a regularization that targets the gradients of the marginal density, aiming to regulate the reliance on non-robust features. Extensive experiments demonstrate the effectiveness of our regularization in boosting model robustness across a wide range of applications.
We emphasize that our approach does not advocate for the complete removal of model reliance on non-robust features, but instead seeks to achieve a balance between model performance and robustness through appropriate regularization strength.

\section*{Acknowledgments}
This research was supported by ARC Discovery Grant 190102443 and by Google Research under Google Research Scholar Program. Dr.~Naveed Akhtar is a recipient of the Australian Research Council Discovery Early Career Researcher Award (project number DE230101058) funded by the Australian Government. Professor Ajmal Mian is the recipient of an Australian Research Council Future Fellowship Award (project number FT210100268) funded by the Australian Government.

\bibliographystyle{splncs04}
\bibliography{main}

\clearpage
\setcounter{page}{1}
\maketitlesupplementary

\section{Proof}\label{app:sec_proof}

In this section, we provide proof of our proposed approach for estimating the gradient difference using the gradients of the output difference.

\begin{proof}[Proof of Equation 7]
Assuming that two functions, $f$ and $g$, are continuously differentiable with respect to $x \in \mathbb R^n$, we can express them using their Taylor series expansions, as
\begin{equation} \label{eq_taylor_1}
f(x) = f(a) + f'(a)(x-a) + o((x-a)^2),
\end{equation}
and
\begin{equation} \label{eq_taylor_2}
g(x) = g(a) + g'(a)(x-a) + o((x-a)^2).
\end{equation}
By subtracting Equation~\ref{eq_taylor_2} from Equation~\ref{eq_taylor_1}, we have
\begin{equation} \label{eq_taylor_3}
f(x)-g(x) = f(a) - g(a) + (f'(a) - g'(a))(x-a) + o((x-a)^2).
\end{equation}
Next, we compute the gradients of both sides of Equation~\ref{eq_taylor_3} with respect to $x$ as
\begin{equation} \label{eq_taylor_4}
\nabla_x (f(x) - g(x)) = f'(a) - g'(a) + o((x-a)^3).
\end{equation}
Then, we can set $x=a$ in Equation~\ref{eq_taylor_4} as
\begin{equation} \label{eq_taylor_5}
\nabla_a (f(a) - g(a)) \approx f'(a) - g'(a).
\end{equation}
Since we have assumed that $f$ and $g$ are differentiable, we can estimate the difference between the gradients of the two functions as the gradient of the difference between the functions as
\begin{equation} \label{eq_taylor_6}
\nabla_x (f(x) - g(x)) \approx \nabla_x f(x) - \nabla_x g(x).
\end{equation}
Since the model $f_{\theta}$ parameterized with $\theta$ is assumed as continuously differentiable, we can substitute the model output logit $f_i(x;\theta)$ and log-softmax output logit $log(\frac{e^{f_i(x;\theta)}}{Z_{f(x)}})$ into the functions $f$ and $g$ in Equation~\ref{eq_taylor_6} as
\begin{equation} \label{eq_taylor_7}
\nabla_x f_i(x;\theta) - \nabla_x log(\frac{e^{f_i(x;\theta)}}{Z_{f(x)}}) \approx  \nabla_x (f_i(x;\theta) - log(\frac{e^{f_i(x;\theta)}}{Z_{f(x)}})).
\end{equation}
Thus, the gradients of the difference between two outputs can be used to approximate the difference between the two gradients of the outputs.
\end{proof}

\section{Norm and Implementation Comparison}\label{app:sec_norm}

In this section, we evaluate the impact of different norms and implementations on our regularizations.

Firstly, we investigate the effect of different $p$-norm values on our regularization approach. Given a variable $p\in \mathbb{R}$, $p$-norm of input $x\in \mathbb{R}^n$ is defined as
\begin{equation} \label{supp_eq_joint}
||x||_p = (|x_1|^p + |x_2|^p + \dots + |x_n|^p)^{1/p}.
\end{equation}

\begin{figure*}[t]
    \begin{subfigure}[c]{0.30\textwidth}
        \centering
        \includegraphics[width=\textwidth]{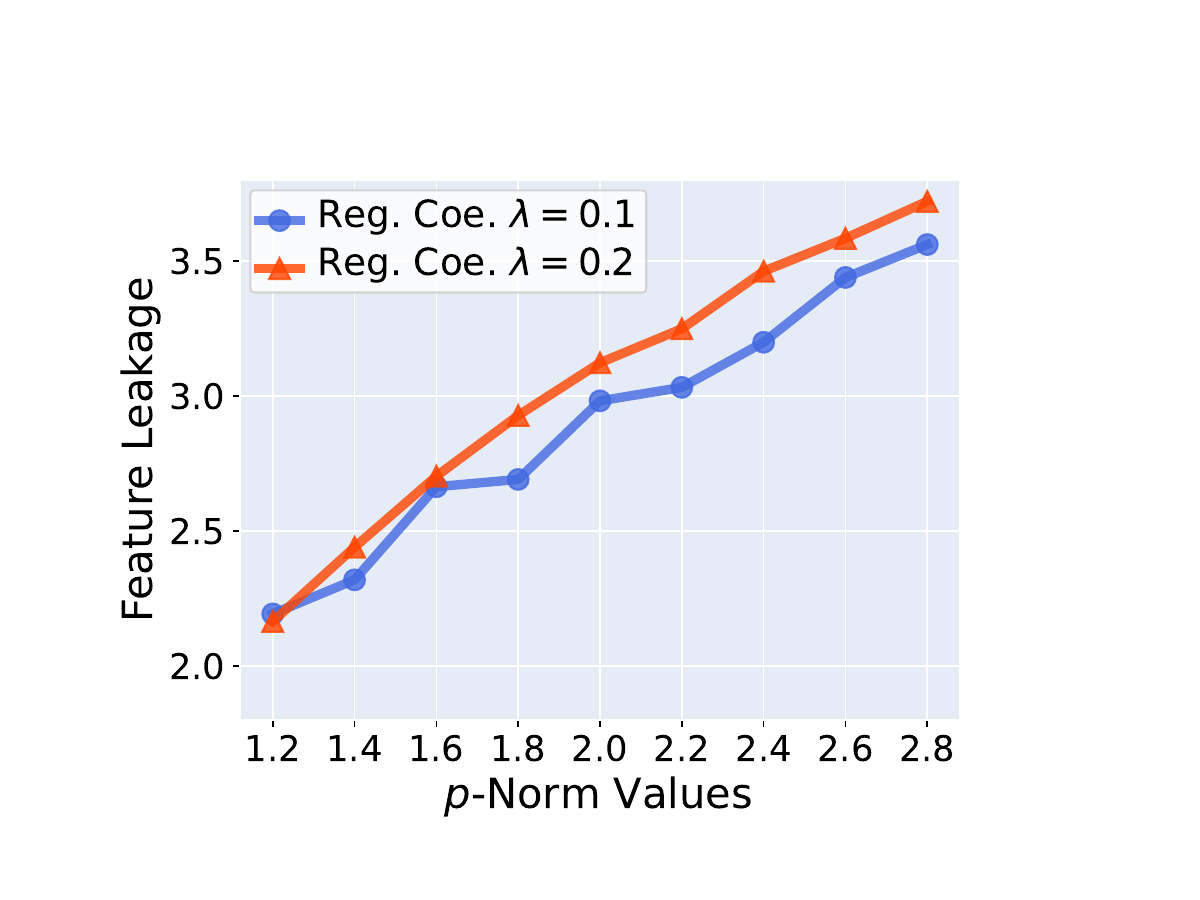}
        \caption{Feature Leakage.}
        \label{fig:norm_leak}
    \end{subfigure}
        \hspace{4mm}
    \begin{subfigure}[c]{0.30\textwidth}
        \centering
        \includegraphics[width=\textwidth]{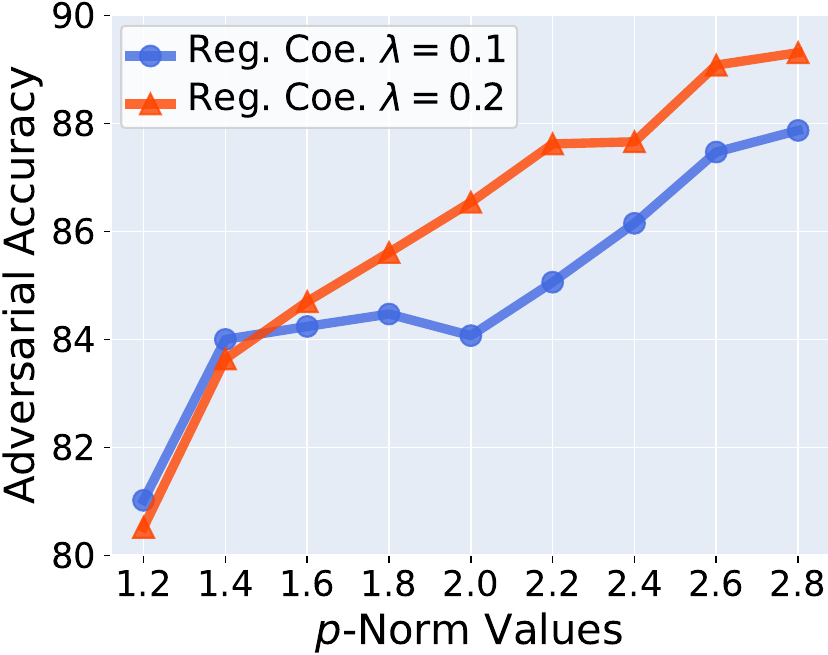}
        \caption{Adv. Accuracy.}
        \label{fig:norm_advacc}
    \end{subfigure}
        \hspace{4mm}
    \begin{subfigure}[c]{0.30\textwidth}
        \centering
        \includegraphics[width=\textwidth]{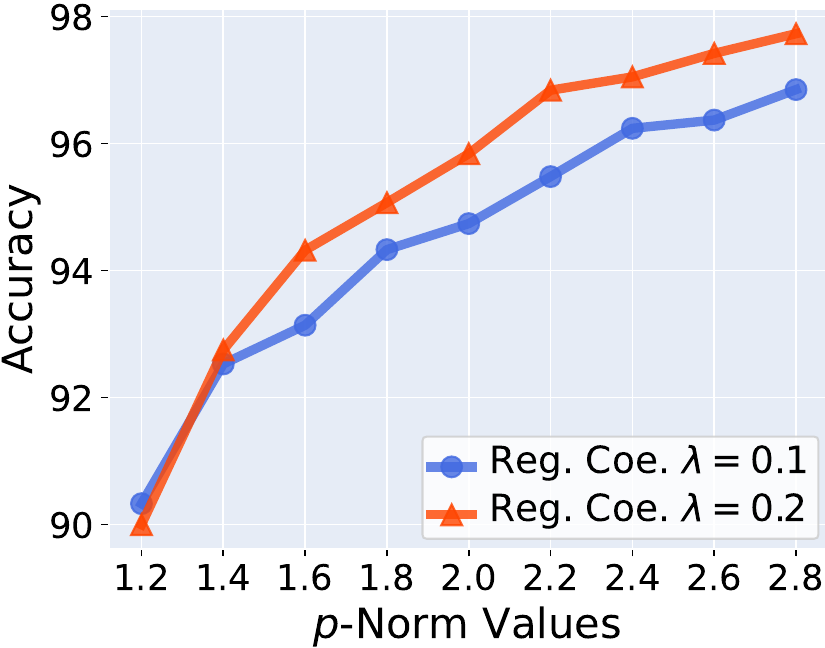}
        \caption{Accuracy.}
        \label{fig:norm_acc}
    \end{subfigure}
    \caption{Results of feature leakage, adversarial accuracy (\%) and standard accuracy (\%) under different $p$-norm values. \textbf{(a)} Smaller $p$-norm values lead to better results in suppressing the feature leakage problem. \textbf{Lower} curves indicate \textbf{better} results. \textbf{(b-c)} Larger $p$-norm values lead to enhanced adversarial robustness and higher accuracy. \textbf{Higher} curves indicate \textbf{better} results.}
\label{fig:norm}
\end{figure*}

The $L_p$ norm allows us to measure the magnitude of a vector using different $p$ values. Different $p$ values exhibit different properties. Smaller $p$ values promote sparsity, while larger $p$ values emphasize the maximum value. Hence, selecting an appropriate $p$ value that strikes a balance between these characteristics is crucial when applying the regularization method to models. In Figure~\ref{fig:norm}, we test the effect of $p$ norm values from $p=1.2$ to $p=2.8$ on models using our regularization with two regularization coefficients $\lambda=0.1$ and $\lambda=0.2$ on BlockMNIST~\cite{Shah2021Do}. In Figure~\ref{fig:norm}(a), lower $p$ values effectively suppress feature leakage, indicating that encouraging sparse features reduces reliance on non-informative features. In Figure~\ref{fig:norm}(b) and Figure~\ref{fig:norm}(c), larger $p$ values lead to enhanced adversarial robustness and higher accuracy, suggesting that models are susceptible to perturbations caused by large gradients. The results reveal that models are easily perturbed from large gradients. The results demonstrate that our regularization enables models to regulate their reliance on non-robust features by adjusting the norm value $p$ and regularization coefficient $\lambda$. In our experiments, we employ $p=2$ for all regularizations to ensure a fair comparison. However, exploring alternative norms in addition to the $p$ norm is expected to further enhance robustness.

\begin{table*}[ht]
\caption{Experimental results on BlockMNIST. Feature leakage, standard accuracy (\%) and adversarial accuracy (\%) under $L_2$ PGD-20 threat models are reported.}
\centering
\label{table:app_robust_BM}
\setlength{\tabcolsep}{1.2mm}{
\begin{tabular}{lccccc}
\toprule[1.5pt]
Method & Feature Leakage $\downarrow$ & Adv. Accu. $\uparrow$ & Accu. $\uparrow$ \\
\hline
MLP & 3.657 & 73.57 & \textbf{99.12}\\
MLP + InputGrad  Reg. & 3.461 & 83.46 & \textbf{94.56}\\
MLP + Our Reg. (Stable \& Efficient) & 2.483 & 84.36 & 94.22\\
MLP + Our Reg. (Stable) & \textbf{2.289} & \textbf{85.71} & 93.45\\
MLP + Our Reg. & \textbf{2.259} & \textbf{85.41} & 93.05\\
\hline
VGG11 & 3.418 & 79.54 & \textbf{99.18}\\
VGG11 + InputGrad  Reg. & 3.792 & 84.72 & 97.35\\
VGG11 + Our Reg. (Stable \& Efficient) & \textbf{2.899} & \textbf{87.79} & \textbf{97.74}\\
VGG11 + Our Reg. (Stable) & \textbf{2.878} & \textbf{84.75} & 96.63\\
\hline
ResNet-18 & 4.113 & 39.46 & \textbf{99.48}\\
ResNet-18 + InputGrad  Reg. & 3.969 & \textbf{89.69} & 97.94\\
ResNet-18 + Our Reg. (Stable \& Efficient) & \textbf{3.408} & 86.67 & 96.29\\
ResNet-18 + Our Reg. (Stable) & \textbf{3.057} & \textbf{89.93} & \textbf{99.16}\\
\bottomrule[1.5pt]
\end{tabular}}
\end{table*}

More experimental results are presented to compare the three implementations of our regularization method on the BlockMNIST dataset. Table~\ref{table:app_robust_BM} shows the results for three different models: MLP, VGG11~\cite{simonyan2014very}, and ResNet-18. Experimental results of models trained with our regularization, including variations with stable and efficient implementations are reported. For MLP models, we observe that both the stable implementation and the efficient implementation of our method achieve outstanding performance compared to the original implementation. This demonstrates the effectiveness of our proposed alternative implementations in enhancing the robustness and performance of MLPs. However, when applied to VGG11 and ResNet-18 models, our regularization compromises their robustness in terms of feature leakage and vulnerability to adversarial perturbations. It can be also observed that ResNet-18, which contains batch normalization (BN) layers~\cite{ioffe2015batch}, exhibits additional performance degradation. This is because BN layers not only introduce the non-linearity operation in the model but also compute gradients with respect to a batch of input samples. This can lead to inaccuracies in the computation of the density gradients. Nevertheless, our implementation still demonstrates robustness compared to the vanilla model and the model using InputGrad regularization. These results suggest that finding a more effective approach to address numerical stability issues and extend the robustness of our regularization method from small to large models is a promising direction for future work.

\section{Adversarial Robustness Comparison}\label{app:sec_adv_rob}
In this section, we present additional results for the comparison of adversarial robustness. Specifically, we evaluate the performance of ResNet-18~\cite{He2016Identity} trained on the CIFAR-100 dataset~\cite{krizhevsky2009learning}, considering both standard accuracy and adversarial accuracy with the varying perturbation budget $\epsilon$. In Table~\ref{table:l2_adv_c100} and Table~\ref{table:linf_adv_c100}, a comprehensive comparison of the adversarial robustness for different models is presented. The first table shows the performance under $L_{2}$ adversarial PGD-20~\cite{Madry2018Towards} attacks, while the second table focuses on the models' performance under $L_{\infty}$ attacks. We compare the standard trained model and three robust models trained with Score-Matching regularization~\cite{srinivas2021rethinking}, InputGrad regularization~\cite{Ancona2018Towards}, and our proposed method. Our results clearly demonstrate the superiority of models trained with our regularization technique. The performance gap is significant, indicating that our approach outperforms the other methods in terms of adversarial robustness under both $L_{\infty}$ and $L_2$ attacks. These experimental findings provide compelling evidence that our regularization technique effectively enhances the model's robustness to adversarial attacks.

\begin{table*}[t]
\caption{Adversarial robustness comparison on ResNet-18. Adversarial accuracy (\%) of the standard trained model (ST) and models with various regularizations under PGD-20 $L_{2}$ attack, along with the standard accuracy (\%), are reported. 
}
\centering
\label{table:l2_adv_c100}
\setlength{\tabcolsep}{1.2mm}{
\begin{tabular}{lcccccc}
\toprule[1.5pt]
Method & Accuracy & $||\epsilon||_2=0.1$ & $||\epsilon||_2=0.3$ & $||\epsilon||_2=0.5$ \\
\hline
ST & 58.62 & 37.40 & 11.54 & 4.92\\
ST + Score-Matching Reg. & 56.66 & 39.78 & 13.88 & 5.29\\
ST + InputGrad Reg. & 57.94 & 40.94 & 14.88 & 5.70\\
ST + Our Reg. & \textbf{58.62} & \textbf{47.03} & \textbf{26.76} & \textbf{15.19}\\
\bottomrule[1.5pt]
\end{tabular}}
\end{table*}

\begin{table*}[t]
\caption{Adversarial robustness comparison on ResNet-18. Adversarial accuracy (\%) of the standard trained model (ST) and models with various regularizations under PGD-20 $L_{\infty}$ attack are reported.}
\centering
\label{table:linf_adv_c100}
\setlength{\tabcolsep}{1.2mm}{
\begin{tabular}{lccccc}
\toprule[1.5pt]
Method & $||\epsilon||_{\infty}=1/255$ & $||\epsilon||_{\infty}=2/255$ & $||\epsilon||_{\infty}=3/255$ \\
\hline
ST & 29.06 & 11.13 & 4.35\\
ST + Score-Matching Reg. & 30.05 & 12.85 & 4.85\\
ST + InputGrad Reg. & 30.95 & 13.49 & 5.16\\
ST + Our Reg. & \textbf{41.10} & \textbf{26.02} & \textbf{13.38}\\
\bottomrule[1.5pt]
\end{tabular}}
\end{table*}

To further confirm generlizability of our method for adversarial robustness, Tab.~\ref{tab_adv} presents additional favorable results on ResNet-18 trained on a subset of Tiny Imagenet~\cite{russakovsky2015imagenet}, SVHN~\cite{netzer2011reading}, and CIFAR-10~\cite{krizhevsky2009learning} against PGD and auto attack (AA)~\cite{Croce2020Reliable}. Moreover, the regularization terms do not increase the parameters, maintaining the same number of parameters as ResNet-18.

\begin{table}[b]
\caption{Results of adversarial robustness on Tiny ImageNet, SVHN and CIFAR-10 against PGD attack and auto attack (AA).}
\centering
\scalebox{1.0}{
\label{tab_adv}
    \setlength{\tabcolsep}{1.2mm}{
    \begin{tabular}{l c c c c c c c c c}
        \toprule[1.5pt]
        \multirow{2}{*}{Method} & 
        \multicolumn{3}{c}{Tiny ImageNet} &
        \multicolumn{3}{c}{CIFAR-10} &
        \multicolumn{3}{c}{SVHN} \\ \cmidrule(r){2-4} \cmidrule(r){5-7} \cmidrule(r){8-10} & 
        Clean & PGD & AA & Clean & PGD & AA & Clean & PGD & AA\\
        \hline
        Score-Mat. & 46.12 & 13.46 & 11.94 & 80.17 & 16.93 & 15.47 & 92.71 & 17.93 & 16.30 \\
        \textbf{Ours} & \textbf{49.58} & \textbf{17.74} & \textbf{16.56} & \textbf{80.86} & \textbf{25.30} & \textbf{24.42} & \textbf{93.08} & \textbf{25.50} & \textbf{24.91} \\
        \bottomrule[1.5pt]
    \end{tabular}}
}
\end{table}

\section{Computational Overhead Analysis}
\begin{figure}[t]
	\centering
	\includegraphics[width=.85\textwidth]{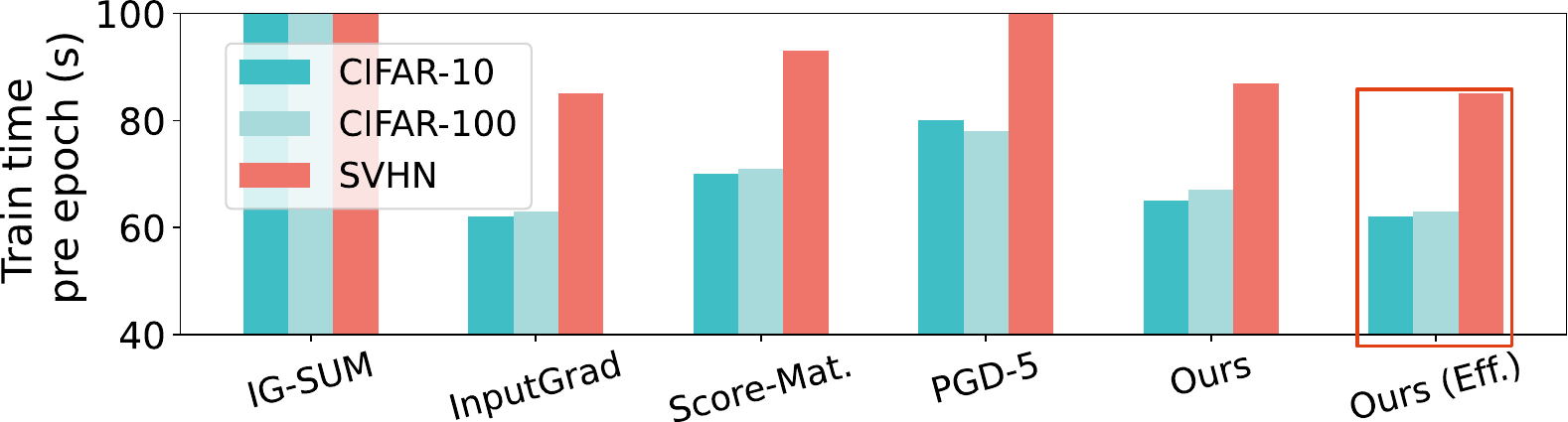}
	\caption{Results of training time comparison between different robustness techniques on CIFAR-10, CIFAR-100 and SVHN datasets.}
	\label{fig_time}
\end{figure}
In this section, we compare the training times of various regularization techniques. Figure~\ref{fig_time} illustrates the results of different robust training methods, including IG-SUM~\cite{Chen2019Robust}, InputGrad~\cite{Ross2018Improving}, Score-Matching~\cite{srinivas2021rethinking}, PGD-5 adversarial training~\cite{Madry2018Towards}, and our proposed regularizations across three datasets: CIFAR-10, CIFAR-100, and SVHN. This efficiency advantage is crucial for large-scale applications where training time can be a bottleneck, making our methods suitable for practical deployment in resource-constrained environments. In summary, our results highlight the effectiveness of our regularizations in achieving a desirable balance between robustness and efficiency.

\section{Robustness against Input Gradient and Density Perturbations}\label{app:sec_pert_rob}

\begin{figure*}[b]
    \hspace{16mm}
    \begin{subfigure}[c]{0.34\textwidth}
        \centering
        \includegraphics[width=\textwidth]{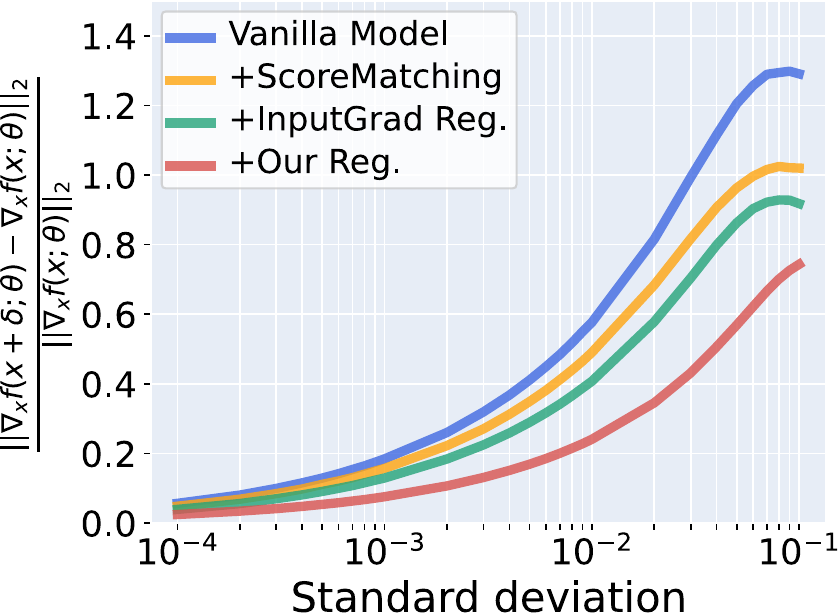}
        \caption{Relative Gradient Robustness.}
        \label{fig:rob_grad}
    \end{subfigure}
    \hspace{4mm}
    \begin{subfigure}[c]{0.34\textwidth}
        \centering
        \includegraphics[width=\textwidth]{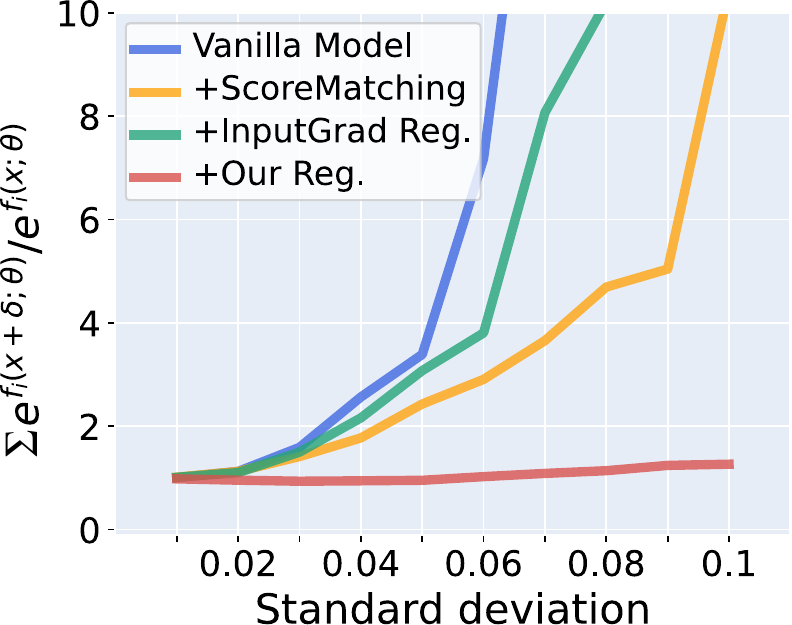}
        \caption{Relative Density Robustness.}
        \label{fig:rob_den}
    \end{subfigure}
    \caption{Robustness comparison against perturbation in input gradients and density on ResNet-18 trained on CIFAR-100. \textbf{(a)}-\textbf{(b)} Relative gradient robustness and relative density robustness against uniform noise with increasing deviation are shown respectively. \textbf{Lower} curves indicate \textbf{better} results.
    }
\label{fig:rob_comp}
\end{figure*}

In this section, we present additional results regarding the robustness of models against Gaussian noise $\delta$ with the increasing standard deviation in both input gradients and density on the CIFAR-100 dataset. We compare the robustness of the vanilla ResNet-18 model and models trained with three regularizations: Score-Matching regularization, InputGrad regularization and our proposed regularization. Figure~\ref{fig:rob_comp}(a) shows the relative gradient robustness.
We can observe that our proposed regularization method not only achieves comparable results but also surpasses other methods by a significant margin.
In Figure~\ref{fig:rob_comp}(b), we present the robustness comparison of the relative density. Notably, our regularization technique leads to a high level of robustness against perturbations in density. The ability to maintain relative density robustness is closely associated with the strong generative capabilities of the models. Moreover, more visualization results by maximizing the activations are provided in Appendix~\ref{sec:act_vis}.

\section{Efficacy for Out-of-Distribution Detection}\label{app:sec_ood}
Out-of-distribution (OOD) detection~\cite{hendrycks2016baseline,liang2017enhancing,grathwohl2020your} is a binary classification problem that aims to identify samples that do not belong to the in-distribution dataset. A robust model is expected to produce discriminative outputs capable of distinguishing between samples from in-distribution and out-of-distribution data. Here, we evaluate the performance of models trained with robust regularizations in detecting OOD samples. To assess OOD detection performance, we follow the recommendation of Hendrycks et al.~\cite{hendrycks2016baseline} and use the area under the receiver-operating curve (AUROC) as the metric.

\begin{table}[b]
\centering
\caption{AUROC results for OOD detection on ResNet-18 models trained on CIFAR-100 and SVHN datasets.}

\label{table:tab_ood}
    \begin{tabular}{l c c c c}
        \toprule[1.5pt]
        \multirow{2}{*}{Method} & 
        \multicolumn{2}{c}{CIFAR-100} &
        \multicolumn{2}{c}{SVHN} \\ \cmidrule(r){2-3} \cmidrule(r){4-5} & 
        $f_i(x;\theta)$ & $f(x;\theta)$ & $f_i(x;\theta)$ & $f(x;\theta)$ \\
        \hline
        Vanilla Model & 0.218 & 0.511 & 0.163 & 0.531 \\
        Score-Matching Reg. & 0.203 & 0.523 & 0.322 & 0.496 \\
        InputGrad Reg. & 0.345 & 0.538 & \textbf{0.419} & 0.570 \\
        \textbf{Our Reg.} & \textbf{0.372} & \textbf{0.507} & 0.339 & \textbf{0.570} \\
        \textbf{Our Reg. (Efficient)} & \textbf{0.378} & \textbf{0.510} & \textbf{0.470} & \textbf{0.590} \\

        \bottomrule[1.5pt]
    \end{tabular}
\end{table}

Table~\ref{table:tab_ood} presents the OOD detection results obtained using ResNet-18 trained on the CIFAR-10 dataset~\cite{krizhevsky2009learning}. In our experiments, we deploy ResNet-18 models trained with different regularizations to detect out-of-distribution samples from both CIFAR-100 and SVHN~\cite{netzer2011reading} datasets. The reported results include the output logits $f(x;\theta)$ and the output logit $f_i(x;\theta)$ of the label $y=i$. Both our regularization and the corresponding efficient implementation results are presented. The outcomes distinctly reveal that our regularization, along with the proposed efficient variant, significantly improves the model's performance in detecting out-of-distribution samples.

\section{Experiments on Numerical Stability Comparison}
\label{app:exp_numerical}
\begin{figure*}[t]
	\includegraphics[width=0.95\textwidth]{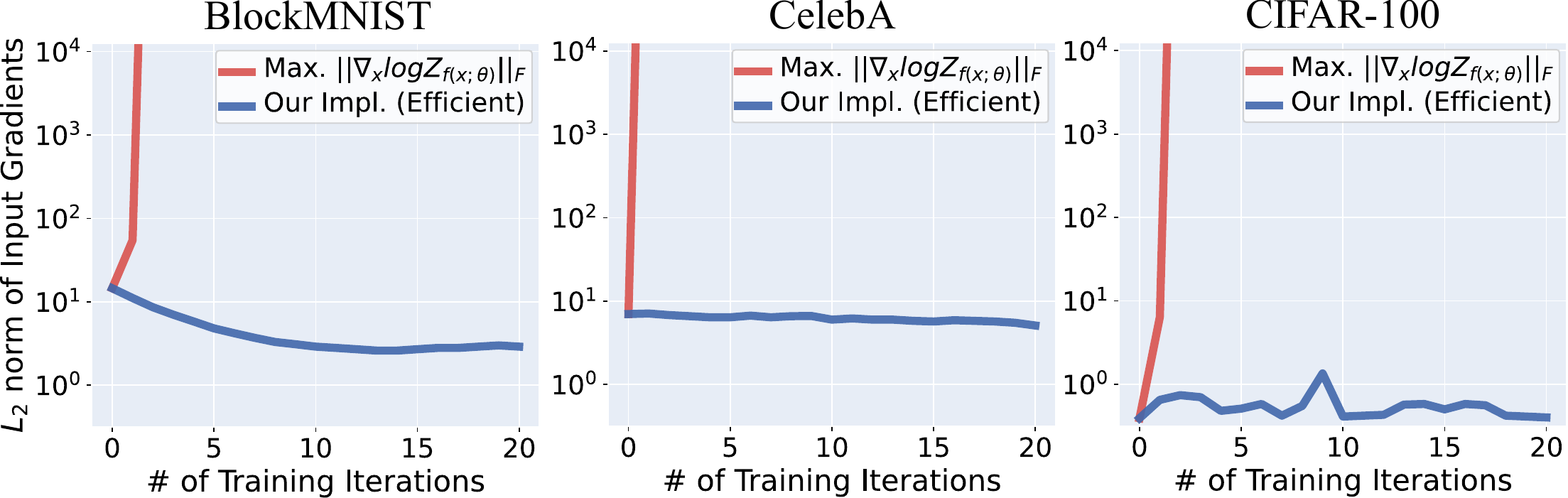}
	\caption{The comparison of numerical stability on BlockMNIST, CelebA and CIFAR-100. The $L_2$ norm of input gradients changes with the number of training iterations.}
	\label{fig:rebb_NS}
\end{figure*}

In this section, we provide more results pertaining to the comparison of numerical stability. Figure~\ref{fig:rebb_NS} illustrates a comprehensive numerical stability analysis conducted on the BlockMNIST, CelebA, and CIFAR-100 datasets. It is observed that the baseline regularization approach exhibits severe numerical instability early in the training phase, whereas our proposed method demonstrates sustained stability throughout the training process.

\section{Activation Visualization}\label{app:sec_act_vis}
\label{sec:act_vis}
\begin{figure*}[t]	\centerline{\includegraphics[width=1.0\textwidth]{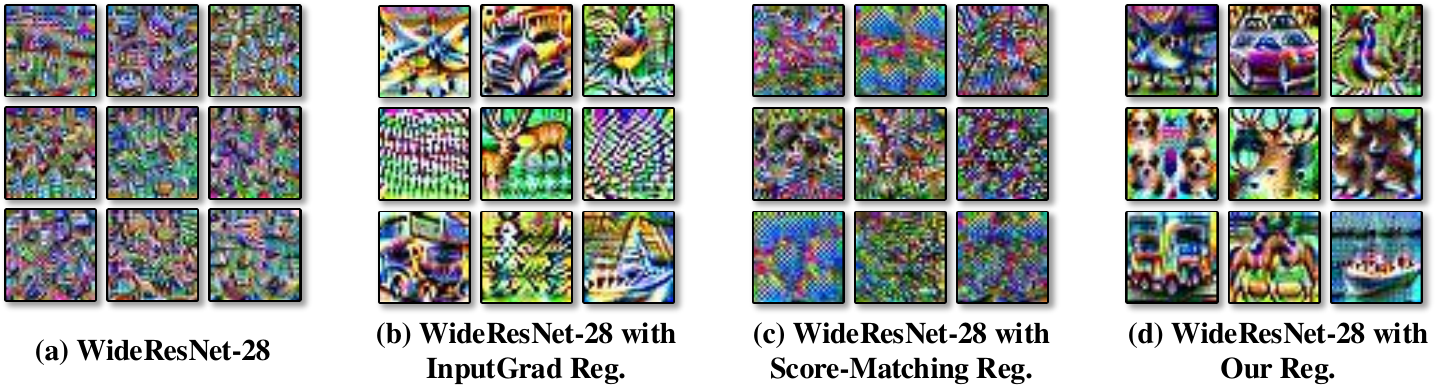}}
	\centering
	\caption{Visualization samples are generated by applying gradient ascent on random inputs using WideResNet-28 trained on CIFAR-10. The visualization results show nine different classes in CIFAR-10. Our method demonstrates superior performance by exhibiting reduced noise and more interpretable patterns in the visualization results.}
	\label{fig:vis_WRN}
\end{figure*}

\begin{figure*}[t]	\centerline{\includegraphics[width=1.0\textwidth]{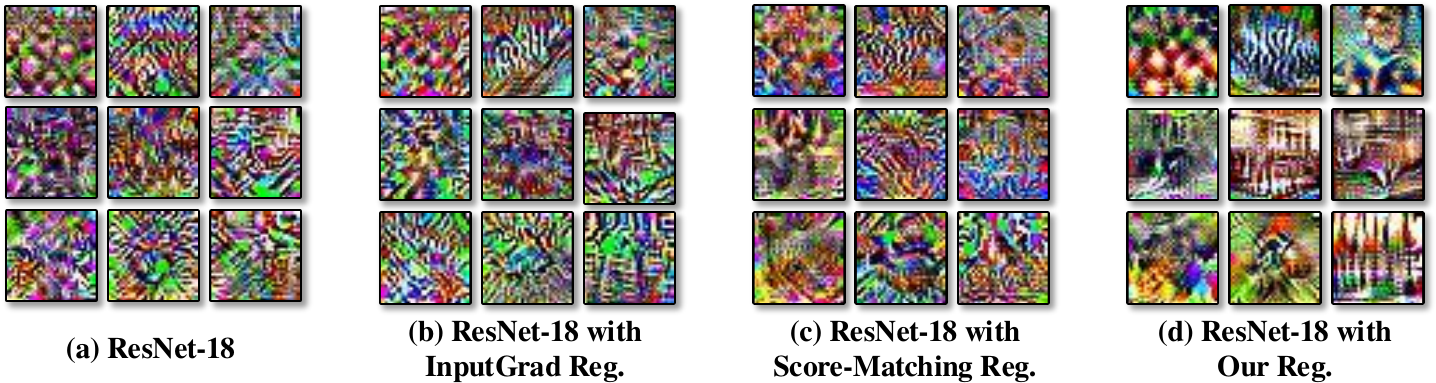}}
	\centering
	\caption{Visualization samples are generated by applying gradient ascent on random inputs using ResNet-18 trained on CIFAR-100. The visualization results show nine different classes in CIFAR-100. Our method demonstrates superior performance by exhibiting reduced noise and more interpretable patterns in the visualization results.}
	\label{fig:vis_RN}
\end{figure*}

In this section, we present visualization samples generated by applying gradient ascent on random inputs. We compare the visualization results of different regularizations on both WideResNet-28~\cite{zagoruyko2016wide} and ResNet-18 models trained on CIFAR-10 and CIFAR-100 respectively. Figure~\ref{fig:vis_WRN} displays the visualization results for WideResNet-28, while Figure~\ref{fig:vis_RN} presents the results for ResNet-18.
The visualization results obtained from the model using our regularization exhibit reduced noise and present more interpretable patterns. The improvement in visualization quality serves as evidence that our technique enhances the interpretability of the underlying features learned by the models.
Our method is effective in enhancing the interpretability and clarity of the learned representations. The reduction of noise and the emergence of more interpretable patterns contribute to a better understanding of the model's decision-making process and aid in capturing relevant features for the respective classes.

\section{Attribution Maps and Insertion Games}\label{app:sec_ins_vis}
\begin{figure*}[t]
	\centerline{\includegraphics[width=1.0\textwidth]{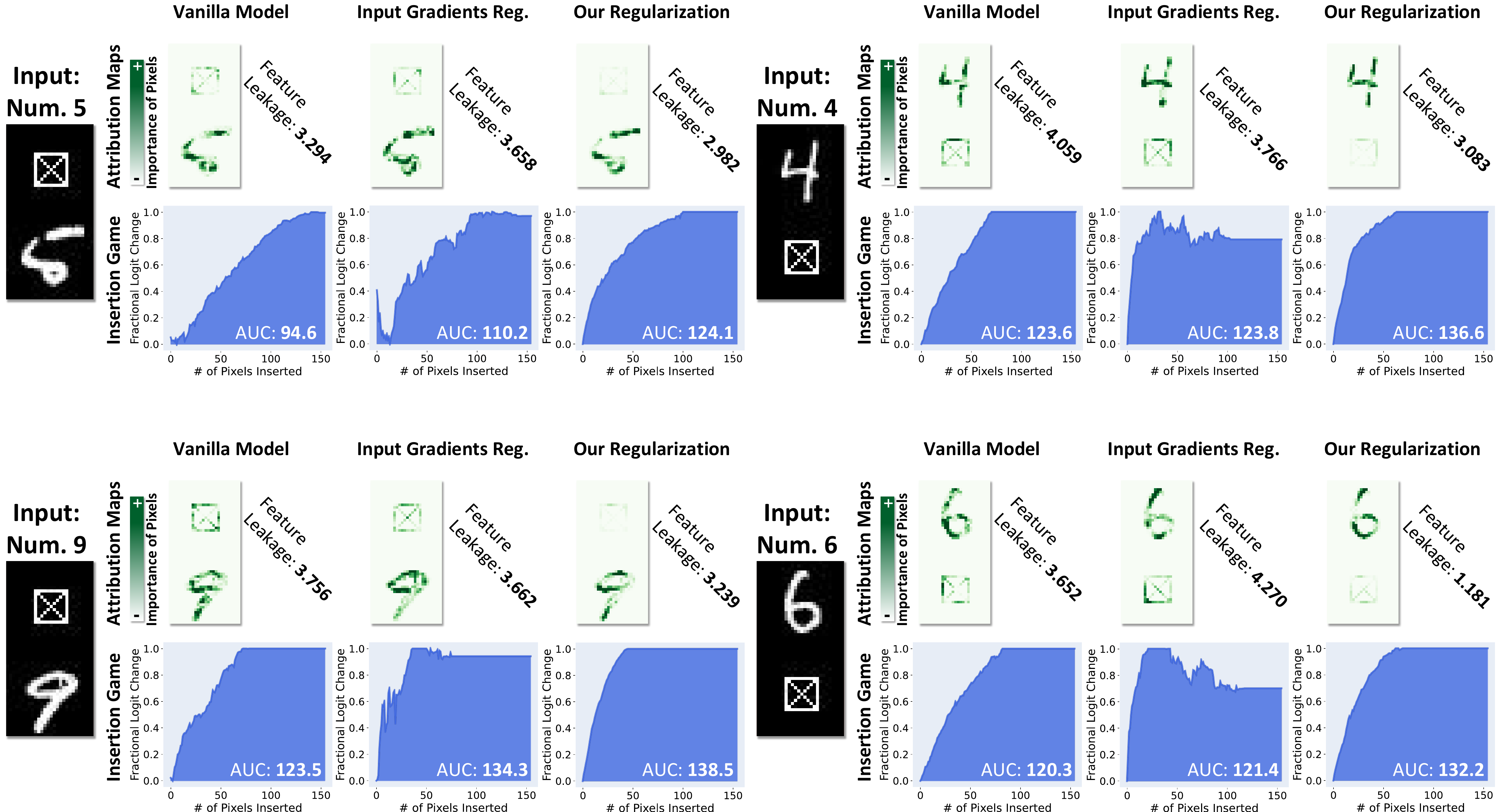}}
	\centering
	\caption{Attribution maps generated by Integrated Gradients and the area under the curve (AUC) of the insertion games for representative samples from BlockMNIST. As compared to the vanilla model and the model with Input Gradient regularization, our regularization leads to interpretable attribution maps with reduced feature leakage and fewer spurious correlations, while also achieving higher AUC for the insertion game.}
	\label{fig:attr_bm}
\end{figure*}

\begin{figure*}[t]
	\centerline{\includegraphics[width=1.0\textwidth]{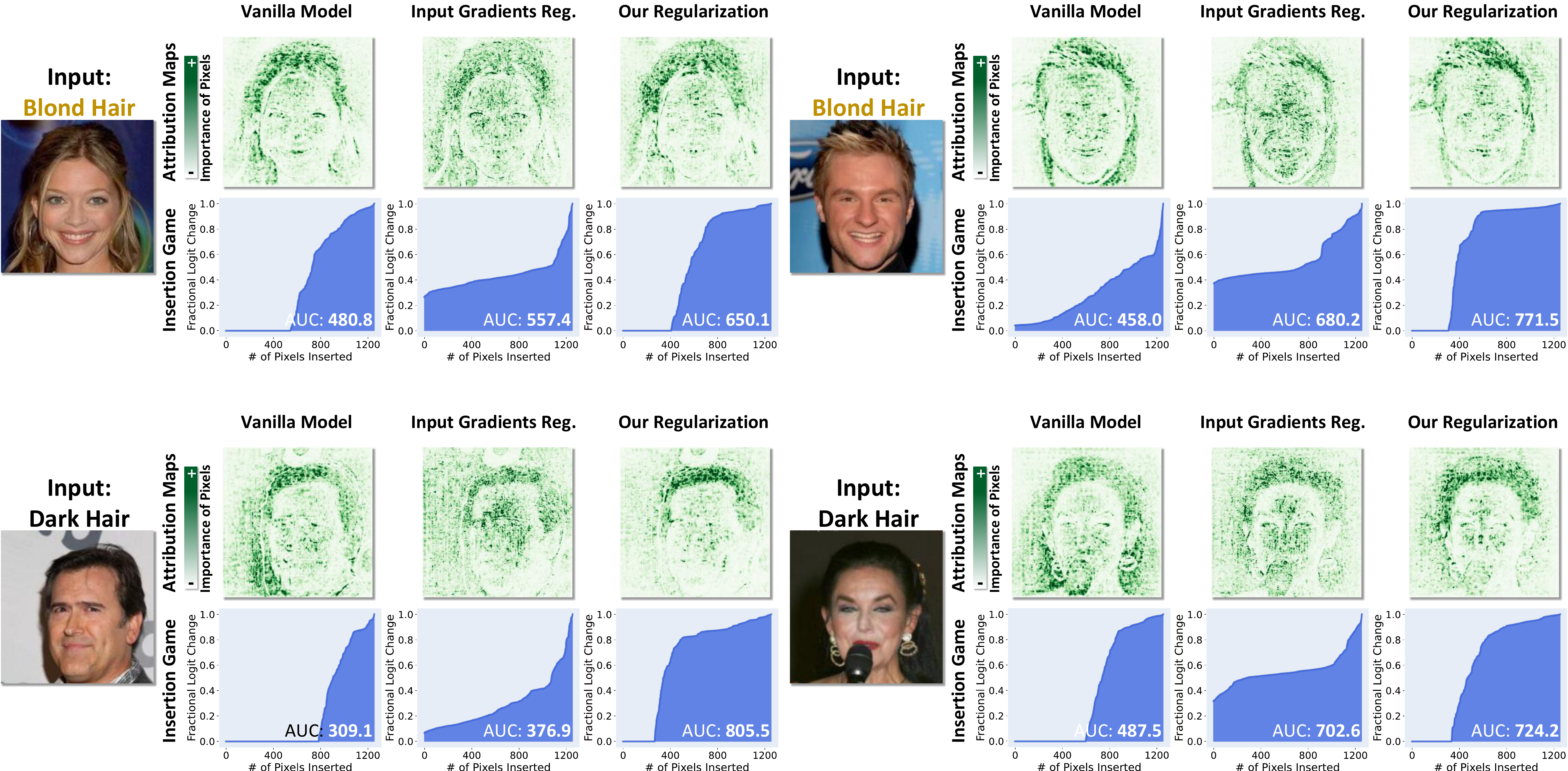}}
	\centering
	\caption{Attribution maps generated by Integrated Gradients and the area under the curve (AUC) of the insertion games for representative samples from CelebA. As compared to the vanilla model and the model with Input Gradient regularization, our regularization leads to lower feature leakage while also achieving higher AUC for the insertion game.}
	\label{fig:attr_celeba}
\end{figure*}

\begin{figure*}[t]
	\centerline{\includegraphics[width=1.0\textwidth]{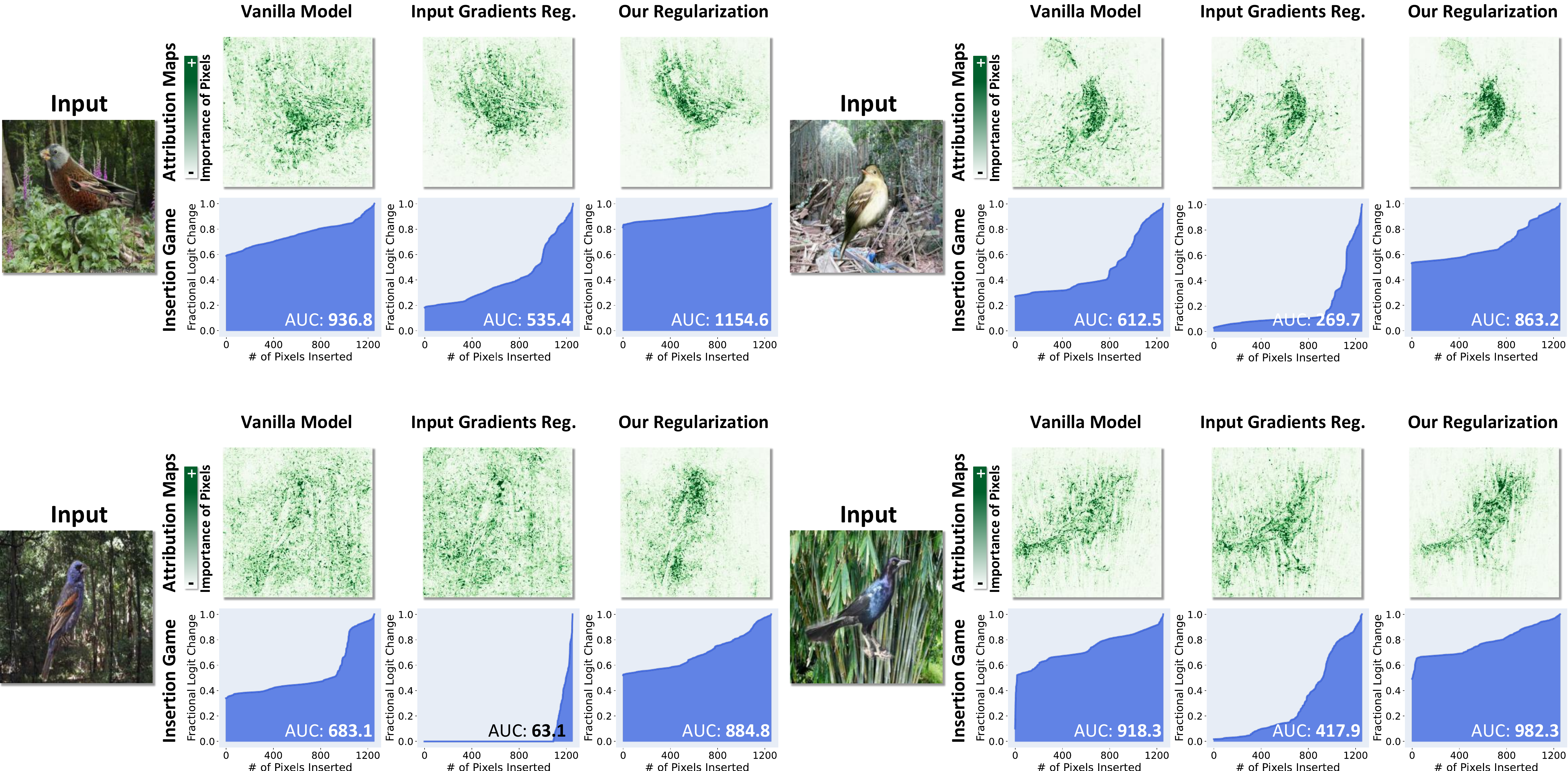}}
	\centering
	\caption{Attribution maps generated by Integrated Gradients and the area under the curve (AUC) of the insertion games for representative samples from Waterbirds. As compared to the vanilla model and the model with Input Gradient regularization, our regularization leads to lower feature leakage while also achieving higher AUC for the insertion game.}
	\label{fig:attr_cub}
\end{figure*}

In this section, we present additional results of attribution maps and insertion games. In Figure~\ref{fig:attr_bm}, Figure~\ref{fig:attr_celeba} and Figures~\ref{fig:attr_cub}, we generate attribution maps using the Integrated Gradients method~\cite{Sundararajan2017Axiomatic} and compute the area under the curve (AUC) of the insertion games for representative samples from BlockMNIST, CelebA and Waterbirds datasets.

The attribution maps demonstrate the effectiveness of our regularization method in suppressing the feature leakage problem on BlockMNIST. The lower values of feature leakage observed in the attribution maps indicate that our approach successfully mitigates the issue of irrelevant or misleading features being attributed to certain classes. Similarly, the attribution maps generated for input from CelebA and Waterbirds dataset show improved interpretability. Moreover, corresponding insertion games are performed to evaluate the model robustness of their highly attributed features. Specifically, the pixels will be interactively inserted in a zero input by their attributions computed by Integrated Gradients. The AUC of the fractional output change with increasing inserted pixels is calculated. For ease of comparison, we sort the output changes of samples in CelebA and Waterbirds datasets.

These results highlight the benefits of our regularization technique, both in terms of improving interpretability and enhancing the model's performance in detecting objects. The results provide additional evidence of the efficacy of our approach in achieving superior performance and robustness.


\section{Limitations}
While evaluation results show our regularization achieves a desirable trade-off, the increased regularization strength can diminish a model's sensitivity to features, compromising the model's overall performance.

In the context of adversarial defense, adversarial training remains preferable for explicitly countering adversarial attacks when computational resources permit. Nonetheless, our approach, which does not depend on specific perturbations, demonstrates high performance across a range of problems, as evidenced by our experimental results.

\section{Experimental Setup}\label{app:sec_exp_setup}
In this section, we provide details regarding the datasets, models, and the experimental platform employed in our experiments.

\subsection{Datasets}
\noindent\textbf{BlockMNIST.} BlockMNIST dataset~\cite{Shah2021Do} is an extension of the MNIST dataset~\cite{Lecun1998Gradient}. Each sample in BlockMNIST is derived from an original MNIST sample by adding a \textit{null} block, which contains non-informative features, randomly positioned at the top or bottom of the image. During the training process, each BlockMNIST sample is generated by randomly attaching the null block to MNIST samples. In the testing process, models are evaluated on the same test samples with fixed-placed null blocks.

\noindent\textbf{CelebA.} 
Liu et al.~\cite{liu2015faceattributes}
introduced the CelebA dataset for facial attribute recognition. Sagawa et al.~\cite{Sagawa2019Distributionally}
further constructed the training set consisting of 162,770 training samples. The smallest group within this dataset comprises male celebrities with blond hair, containing 1,387 samples. In our experiment, we adopt the same dataset configuration, with hair color (blond \& dark) as the target attribute and gender (male \& female) as the spurious correlated features.

\noindent\textbf{Waterbirds.} Waterbirds dataset~\cite{Sagawa2019Distributionally} is constructed by combining the CUB-200-2011~\cite{wah2011caltech} and Places datasets~\cite{zhou2017places}. Specifically, the bird images from CUB-200-2011 are cropped using segmentation annotations and then positioned on backgrounds from the Places dataset, which consists of land or water scenes. The placement of the bird images on the backgrounds is determined by the category of the birds, i.e., whether they are land or water birds. Consistent with the settings in~\cite{Sagawa2019Distributionally}, we follow the same approach of placing 95\% of all waterbirds against a water background and the remaining 5\% against a land background.

\noindent\textbf{CIFAR-10 and CIFAR-100.} CIFAR-10 and CIFAR-100 datasets~\cite{krizhevsky2009learning} are widely utilized for evaluating the recognition capabilities of various models. The CIFAR-10 dataset consists of 60,000 images, each with dimensions of 32$\times$32$\times$3, and is divided into 10 different classes, with 6,000 images per class. The dataset is further partitioned into a training set containing 50,000 samples and a separate test set comprising 10,000 samples. Similarly, CIFAR-100 also comprises 60,000 images, but it offers a more fine-grained classification task with 100 distinct classes.

\noindent\textbf{SVHN.} SVHN dataset~\cite{netzer2011reading} is a collection of real-world images depicting house numbers captured from street views. It contains a training set of 73,257 images and a test set of 26,032 images. The dataset exhibits diverse variations in lighting conditions, viewpoints, and digit appearances, reflecting the challenges encountered in real-world scenarios.

\subsection{Models}
\noindent\textbf{MLP.}
We train two-hidden-layer MLPs using different techniques and regularizations on the BlockMNIST dataset for 80 epochs with a learning rate of 0.0001. In our experiments, two $L_{\infty}$ adversarially trained models are compared. To augment the training samples, we generated perturbations using the PGD~\cite{Madry2018Towards} attack. For PGD adversarial accuracy, we test all models under $L_{\infty}$ and $L_{2}$ threats with steps of $\alpha=0.01$, within the perturbation budgets of $\epsilon=0.3$ followed by Tsipras~\etal ~\cite{Tsipras2019Robustness}. 

\noindent\textbf{ResNet.}
In CelebA and Waterbirds datasets, we train ResNet-34 models~\cite{He2016Identity} with different regularizations for 50 epochs and 300 epochs separately. Vanilla ResNet-34 and ResNet-34 with GroupDRO~\cite{Sagawa2019Distributionally} are trained with a learning rate of 0.0001 and all compared models with different regularizations are trained with a learning rate decayed by 10. In the training process, each batch of training samples is re-weighted to have the same number of samples in each group. In CIFAR-10 and CIFAR-100 dataset, we train ResNet-18 and ResNet-34 for 200 epochs with a learning rate of 0.01 decayed by 10 in the 100-th and 175-th epochs. 

\noindent\textbf{WideResNet.}
To perform activation visualization, WideResNet-28~\cite{zagoruyko2016wide} is also trained in CIFAR-10 for 200 epochs with a learning rate of 0.01 decayed by 10 in the 100-th and 175-th epochs.

\noindent\textbf{VGGNet.} We trained VGG11 models~\cite{simonyan2014very} on the BlockMNIST dataset using various techniques and regularizations. The training process involved 80 epochs with a fixed learning rate of 0.0001.

\subsection{Experimental Platform}
All experiments were performed on a Linux machine equipped with an NVIDIA GTX 3090Ti GPU featuring 24GB of memory. The machine also consisted of a 16-core 3.9GHz Intel Core i9-12900K CPU and 128GB of main memory. The models were tested and trained using the PyTorch deep learning framework (v1.12.1) in the Python programming language.

\end{document}